\documentclass{article}

\PassOptionsToPackage{numbers, compress}{natbib}

\usepackage[final]{neurips_2024}
\usepackage{algorithm,algorithmic,fancyhdr,times,wrapfig}

\usepackage{amsmath,amsfonts,bm}

\def\eqref#1{equation~\ref{#1}}

\def\1{\bm{1}}

\DeclareMathAlphabet{\mathsfit}{\encodingdefault}{\sfdefault}{m}{sl}
\SetMathAlphabet{\mathsfit}{bold}{\encodingdefault}{\sfdefault}{bx}{n}

\def\gD{{\mathcal{D}}}
\def\gE{{\mathcal{E}}}

\def\gM{{\mathcal{M}}}

\def\gZ{{\mathcal{Z}}}

\newcommand{\E}{\mathbb{E}}

\newcommand{\R}{\mathbb{R}}

\DeclareMathOperator*{\argmin}{arg\,min}

\usepackage[toc,page,header]{appendix}
\usepackage{minitoc}

\usepackage[utf8]{inputenc} %
\usepackage[T1]{fontenc}    %
\usepackage{nicefrac}       %
\usepackage{microtype}      %
\usepackage{xcolor}         %

\usepackage{paralist}
\usepackage{float}
\usepackage{enumerate}
\usepackage{enumitem}
\usepackage[ruled,noend,algo2e]{algorithm2e}
\usepackage[breaklinks=true,pagebackref=true]{hyperref}
\usepackage{footnotebackref}
\usepackage{url}
\usepackage{subcaption}
\usepackage{graphicx}
\graphicspath{ {./Plots/} }
\usepackage{amsfonts,xspace}
\usepackage{lmodern}
\usepackage{siunitx}
\usepackage{booktabs}
\usepackage{etoolbox}
\usepackage{comment}
\SetKwInOut{Input}{input}
\SetKwInOut{Output}{output}
\SetKwComment{Comment}{// }{}

\usepackage[capitalize,nameinlink]{cleveref}
\crefformat{equation}{\textup{#2(#1)#3}}
\crefrangeformat{equation}{\textup{#3(#1)#4--#5(#2)#6}}
\crefmultiformat{equation}{\textup{#2(#1)#3}}{ and \textup{#2(#1)#3}}
{, \textup{#2(#1)#3}}{, and \textup{#2(#1)#3}}
\crefrangemultiformat{equation}{\textup{#3(#1)#4--#5(#2)#6}}%
{ and \textup{#3(#1)#4--#5(#2)#6}}{, \textup{#3(#1)#4--#5(#2)#6}}{, and \textup{#3(#1)#4--#5(#2)#6}}

\Crefformat{equation}{#2Equation~\textup{(#1)}#3}
\Crefrangeformat{equation}{Equations~\textup{#3(#1)#4--#5(#2)#6}}
\Crefmultiformat{equation}{Equations~\textup{#2(#1)#3}}{ and \textup{#2(#1)#3}}
{, \textup{#2(#1)#3}}{, and \textup{#2(#1)#3}}
\Crefrangemultiformat{equation}{Equations~\textup{#3(#1)#4--#5(#2)#6}}%
{ and \textup{#3(#1)#4--#5(#2)#6}}{, \textup{#3(#1)#4--#5(#2)#6}}{, and \textup{#3(#1)#4--#5(#2)#6}}

\crefdefaultlabelformat{#2\textup{#1}#3}
\usepackage{amsmath,amsthm,mathtools,amssymb}
\makeatletter
\newtheorem*{rep@theorem}{\rep@title}
\newcommand{\newreptheorem}[2]{%
\newenvironment{rep#1}[1]{%
 \def\rep@title{#2 \ref{##1}}%
 \begin{rep@theorem}}%
 {\end{rep@theorem}}}
\makeatother

\newtheorem{definition}{Definition}

\newreptheorem{lemma}{Lemma}

\newtheorem{theorem}{Theorem}
\newreptheorem{theorem}{Theorem}
\crefname{assumption}{Assumption}{Assumptions}

\newcommand{\dist}{\mbox{\rm dist}}

\DeclareMathOperator{\diag}{diag}
\DeclareMathOperator{\Diag}{Diag}
\newcommand*{\lmss}{\fontfamily{lmss}\selectfont}
\newcommand{\norm}[1]{{\left\|{#1}\right\|}}
\newcommand{\prox}{\mbox{\rm prox}}
\def\pg{{\hyperref[alg:solver]{\lmss PG}}\xspace}
\def\msgd{{\lmss MSGD}\xspace}
\def\adam{{\lmss Adam}\xspace}
\def\adamw{{\lmss AdamW}\xspace}
\def\mda{{\lmss MDA}\xspace}
\def\madgrad{{\lmss MADGRAD}\xspace}
\def\rmda{{\lmss RMDA}\xspace}
\def\ramda{{\lmss RAMDA}\xspace}
\def\proxsgd{{\lmss ProxSGD}\xspace}
\def\proxssi{{\lmss ProxSSI}\xspace}
\def\proxgen{{\lmss ProxGen}\xspace}

\newcommand{\inprod}[2]{\langle#1,\,#2\rangle}
\newcommand{\as}{\xrightarrow{\text{a.s.}}}

\title{Regularized Adaptive Momentum Dual Averaging with an
Efficient Inexact Subproblem Solver for Training Structured Neural
Network}

\author{%
  Zih-Syuan Huang\\ 
  Department of Computer Science and Information Engineering\\ 
  National Taiwan University\\
  Taipei 106, Taiwan\\
  \texttt{r11922210@ntu.edu.tw}
  \And
  Ching-pei Lee \\ 
  Department of Advanced Data Science\\
  Institute of Statistical Mathematics\\
  Tachikawa, Tokyo 190-8562, Japan\\
  \texttt{chingpei@ism.ac.jp}
}

\begin{document}
\doparttoc %
\faketableofcontents %

\maketitle

\begin{abstract}
We propose a Regularized Adaptive Momentum Dual Averaging
(\ramda) algorithm for training structured neural networks.
Similar to existing regularized adaptive methods, the subproblem
for computing the update direction of \ramda involves a nonsmooth
regularizer and a diagonal preconditioner, and therefore
does not possess a closed-form solution in general.
We thus also carefully devise an implementable inexactness
condition that retains convergence guarantees similar to the exact
versions, and propose a companion efficient solver for the
subproblems of both \ramda and existing methods to make them
practically feasible.
We leverage the theory of manifold identification in variational
analysis to show that, even in the presence of such inexactness,
the iterates of \ramda attain the ideal structure induced by the
regularizer at the stationary point of asymptotic convergence.
This structure is locally optimal near the point of convergence,
so \ramda is guaranteed to obtain the best structure possible
among all methods converging to the same point,
making it the first regularized adaptive method outputting models
that possess outstanding predictive performance while being
(locally) optimally structured.
Extensive numerical experiments in large-scale modern computer
vision, language modeling, and speech tasks show that the proposed
\ramda is efficient and consistently outperforms state of the art
for training structured neural network.
Implementation of our algorithm is available at 
\url{https://www.github.com/ismoptgroup/RAMDA/}.
\end{abstract}

\section{Introduction}
\label{sec:intro}
Since the recent emergence of ChatGPT, large language models (LLMs)
and other huge deep learning models have garnered much attention and
popularity, even among the public who are unfamiliar with machine
learning.
A challenge with such gigantic neural network models is their vast
number of model parameters, reaching hundreds of billions, resulting
in expensive storage and inference.
It thus becomes crucial to find ways to exploit structures in trained
models to reduce their spatial and prediction costs without degrading
the prediction performance.
An active line of research is to explicitly add a nonsmooth
regularization term to the training objective function and apply
proximal stochastic (sub)gradient methods,
with or without a diagonal
	preconditioner for adaptiveness, to induce a pre-specified type of
	desirable structure in the final model
	\citep{yang2019proxsgd,JY20a,deleu2021structured}.
Unfortunately, although the added regularizer indeed induces some desirable
structures at the stationary points of the training objective function,
the iterates of these methods only converge to those stationary points
asymptotically, but never really attain such a point at any iteration.
Therefore, whether the output model of these algorithms, which is also
an iterate that is only close enough to a stationary point, indeed
possesses the ideal structure at the nearby stationary point is
unknown, and theoretical analyses of these algorithms do not cover any
guarantees regarding the obtained structure.
Indeed, \cite{ZSH21a} oberserd empirically that the structures
obtained by those methods are highly suboptimal and unstable over
iterations.
They then proposed a regularized dual averaging method called \rmda,
and
proved that after a finite number of steps, the iterates of \rmda can
stably identify the locally optimal structure induced by the
regularizer at the stationary point of asymptotic
convergence.\footnote{See the first paragraph in Section~1 and
	Appendix~B of \cite{ZSH21a} for a discussion about why the
structure at the point of convergence is locally optimal.}
This is up to our knowledge the only method with such
structure guarantees for training structured neural networks.
With this property, their experiments demonstrated that their method
also empirically outperforms existing methods on modern computer
vision tasks.
However, since \rmda does not incorporate adaptiveness and their
experiments are conducted only on medium-scale image classification
problems, its usefulness beyond computer vision is in doubt.

For a wide range of tasks in deep learning such as language modeling
and speech recognition,
researchers have developed numerous architectures to achieve
state-of-the-art prediction performance, including
the transformer \citep{AV17a} and the LSTM \citep{SH97a}.
The transformer is also gaining prominence
in computer vision for achieving exceptional performance
\citep{Liu21a}.
Therefore, it is becoming increasingly important to devise
methods that attain satisfactory performance for training these
network architectures with structure.
For such modern architectures, adaptive methods like \adam
\citep{DPK15a} that iteratively rescale the stochastic gradient update
directions via a coordinate-wise/diagonal preconditioner are known to
outperform their non-adaptive counterparts and thus considered
state-of-the-art \citep{Den17a, Anil19a, JZ20a, Liu20a, Kun23a}.
It is hence expected that the non-adaptive \rmda of \cite{ZSH21a} might not
lead to promising results for such widely-used architectures and tasks.

This work aims to fill this gap to propose a practical regularized
adaptive method with guarantees for both convergence and structure
identification.
Since \rmda already has structure guarantees, it might look like we
just need to combine it with an arbitrary preconditioner for
adaptiveness.
However, this seemingly easy extension actually requires deliberation
in two aspects.
First, except for few exceptions, combination of even a simple
diagonal preconditioner and a nonsmooth regularizer makes the
training subproblem complicated with no closed-form solution.
This is totally different from adaptive methods with no regularization,
whose subproblem optimal solution can be easily computed by
coordinate-wise divisions.
Therefore, in the regularized case, the best we can hope for is to
apply an iterative approach to approximately solve the subproblem.
This calls for careful design and control for the measure and the
degree of the inexactness in the approximate subproblem solution.
The second aspect is the need of an appropriate preconditioner that
provides not only outstanding empirical performance but also desirable
theoretical properties.
The interplay between the inexactness and the preconditioner makes it
particularly difficult to address the following three challenges
simultaneously.
(i) \emph{Convergence}: Proving convergence of a new algorithm with even
just one added component is always a nontrivial task.
For example, although convergence of SGD has been well-studied
for decades, similar guarantees for its adaptive correspondence,
Adagrad, is not established until very recently \citep{DefBBU22a}.
We are dealing with both a preconditioner that changes the whole
algorithm, just like from SGD to Adagrad, and the inevitable inexact
subproblem solutions that could nullify many useful properties
(regarding the subdifferential) commonly used in convergence proofs.
(ii) \emph{Structure}: Theoretical guarantees for structure
identification is another critical aim of this work.
Inexactness alone already makes this goal difficult;
see Example~1 of \cite{Lee23a} for a simple instance such that even
infinitesimal inexactness could hinder structure identification.
Even without inexactness, finding a preconditioner that leads to
structure identification guarantees is already difficult because no
adaptive algorithm, even in the much simpler deterministic and exact
setting, is known to have such a guarantee.
(iii) \emph{Subproblem solver}: Our goal is a practical algorithm,
so we need to solve the subproblem efficiently.
This requires the inexact measure be checkable and the degree quickly
attainable by a well-designed solver, and the preconditioner
should make the subproblem well-conditioned and cannot complicate the
computation of the solver.

To tackle these difficulties, we start from considering structure
identification.
We
leverage the theory of manifold identification in variational analysis
and nonlinear optimization
to design a method that leads to finite-iteration structure
identification guarantees.
As discussed by \cite{PooLS18a,ZSH21a}, the key to such guarantees for
stochastic algorithms is to ensure the variance in the stochastic
estimations decreases to zero.
Due to the standard practice of data augmentation in deep learning,
the training loss in the objective function is essentially the
expected value of the training loss over a certain probability
distribution instead of a finite sum.
We thus draw inspirations from \cite{LeeW12a,ZSH21a} to consider
a dual-averaging-type approach
\citep{nesterov2009primal,LX10a} with momentum to attain
variance reduction in this setting for the stochastic gradient estimation.
However, we also need variance reduction for the preconditioner,
so we carefully select a preconditioner whose update is in a manner
similar to dual averaging, and prove that its variance also decreases
to zero.
We then conceive an implementable and practical subgradient-norm-based
inexactness measure compatible with the structure identification
theory.
Further requirements are then added to the inexactness degree and the
preconditioner to ensure convergence, and we also safeguard the
preconditioner to keep the subproblems well-conditioned and the
computation simple.
We then propose to solve the subproblem by a proximal gradient (PG)
solver that provably achieves our inexactness requirement efficiently.
This leads to our Regularized Adaptive Momentum Dual Averaging
(\ramda) algorithm.

We summarize our main contributions as follows.
\begin{enumerate}[leftmargin=*,topsep=-2pt]
	\item \textbf{\emph{An adaptive algorithm for finding locally optimal
		structures}}:
		\ramda is the first regularized \emph{adaptive} method guaranteed
		to find the locally optimal structure possessed by the
		stationary point to which its iterates converge.
		It thus produces models that are more structured while
		retaining the superb prediction performance of adaptive methods.

	\item \textbf{\emph{Efficient subproblem solver for regularized
		adaptive methods}}:
		We propose an implementable inexactness condition and
		a companion efficient subproblem solver for regularized
		adaptive methods (including ours and existing ones) whose
		subproblems have no closed-form solution.
		We show that the induced inexactness does not affect
		convergence or structure identification guarantees.
		This condition and subproblem solver thus also serve as a key step
		for realizing existing frameworks for regularized adaptive methods.

\item \textbf{\emph{A method with outstanding empirical performance}}:
Experiments on training modern neural networks in computer vision
(ImageNet), language modeling (Transformer-XL), and speech
(Tacotron2) with structured sparsity show that \ramda steadily
outperforms state of the art by achieving higher structured sparsity
ratio and better prediction performance simultaneously.
\end{enumerate}

\section{Related Work}
\label{sec:related}
\paragraph{Dual Averaging for Deep Learning.}
Our method is motivated by \cite{ZSH21a} that adapted the famous
regularized dual averaging \citep{LX10a,LeeW12a}
approach with momentum to train structured neural network models
with data augmentation.
They selected dual averaging for the gradient estimation to achieve
variance reduction for structure guarantees, but their algorithm does
not allow for adaptiveness.
Inspired by this approach, we also take dual-averaging-like updates
for the diagonal preconditioner in the subproblem for
adaptiveness.
Our preconditioner design also borrows ideas from the empirically
successful \madgrad of \cite{DA21a} for training non-regularized
neural networks.
\ramda can thus also be seen as a generalization of \madgrad to the
regularized setting.
Since no regularizer is present, unlike \ramda, the subproblem of
\madgrad has a closed-form solution and no structure is expected.
Moreover, \cite{DA21a} only analyzed convergence rates of the
objective value when the problem is convex.
Our analysis of (i) variance reduction in the preconditioner,
(ii) convergence in the nonconvex nonsmooth regularized case,
and (iii) structure identification guarantees are novel and closer to
properties desirable in practice.
The first two items are also applicable when no regularizer is
present, so our theory also expands guarantees for \madgrad.

\paragraph{Regularized Stochastic Algorithms for Deep Learning.}
Other than \rmda, there are several works on training
structured neural networks through regularization and its proximal
operator, but none have structure guarantees.
\cite{yang2019proxsgd} considered a simple regularized SGD
method with momentum, but their convergence analysis is only for
the nonadaptive case.
\cite{JY20a} studied a general regularized adaptive framework
\proxgen that incorporates diagonal preconditioners, and showed
that the subgradient of the objective function can decrease
to the reciprocal of the batch size, but their result does not
guarantee further convergence to stationary points.
Moreover, they do not allow inexactness in the subproblem,
so their framework can be realized for only a small class of problems.
\cite{deleu2021structured} proposed \proxssi that extends \proxgen to
the case of group-sparsity regularizers, whose corresponding
subproblem indeed has no closed-form solution.
They applied the Newton-Raphson method to obtain
nearly-optimal subproblem solutions,
and proposed a seemingly mild inexactness condition.
Unfortunately, their condition is not checkable,
and their corresponding convergence guarantee requires the regularizer
to be locally smooth around each iterate, which excludes most
regularizers that induce meaningful structures.
On the other hand, we will show that with our implementable
inexactness condition,
\proxgen still possesses the same convergence guarantees in \cite{JY20a}
without any additional requirement on the regularizer.
Moreover, we will see in \cref{sec:exp} that the time cost of the
subproblem solver of \proxssi is prohibitively high.

\paragraph{Structure and Manifold Identification.}
The major tool for our structure guarantees is the theory of
manifold identification \citep{HarL04a,HarL07a,LewZ13a,Lee23a} in variational
analysis and nonlinear optimization.
This theory shows that points possessing the same structure induced by
the regularizer at a stationary point form a smooth manifold around
this stationary point,
and with properties from the regularizer, if a sequence of points
converges to this stationary point with their corresponding
subgradients decreasing to zero,
this sequence is guaranteed to eventually stay in this manifold, thus
identifying the structure.
\cite{LeeW12a,PooLS18a,ZSH21a} have leveraged this tool
to show manifold identification for various stochastic algorithms, and
the common key, as pointed out by \cite{PooLS18a}, is variance
reduction. Our analysis uses a result given in \cite{Rus80a} to
prove so for both the gradient estimator and the preconditioner.

\section{Problem Setting and Algorithm}
\label{sec:alg}
As described in \cref{sec:intro}, we consider the case in which the training objective
function is the expectation over a probability distribution as
follows.
\begin{equation}
\min\nolimits_{W \in \gE}\quad F\left( W \right) \coloneqq \E_{\xi\sim \gD}
\left[ f_{\xi} \left( W \right) \right] + \psi\left( W \right),
\label{eq:f}
\end{equation}
where $\gE$ is a Euclidean space with inner product
$\inprod{\cdot}{\cdot}$ and its induced norm $\norm{\cdot}$,  $\gD$
is a distribution over a space $\Omega$ representing all possible
data modifications, $f_\xi$ is differentiable almost everywhere for
any $\xi$, and the possibly nonsmooth regularizer $\psi(W)$ is for
promoting a desirable structure in the optimal solutions.

Our algorithm can be seen as a double-dual averaging method that
incorporates momentum, a proximal operation for the regularization,
and dual averaging for updating both the stochastic gradient
estimation and the preconditioner.
For ease of description, we assume without loss of generality
that $\gE = \R^n$ in this section.
At the $t$th iteration with learning rate $\eta_t$ and iterate
$W^{t-1}$, we first draw an independent and identically distributed
sample $\xi_t \sim \gD$,
compute the stochastic
(sub)gradient $G^t \coloneqq \nabla f_{\xi_t}(W^{t-1})$ of the loss function at the current
point $W^{t-1}$ with respect to $\xi_t$,
and then update the weighted sum $V_t$ of historical stochastic
gradients and the weighted sum $U_t$ of their squared norms
using the value $s_t$:
\begin{equation}
	\label{eq:da}
	\begin{cases}
		V_0 \coloneqq 0,\quad V_t \coloneqq V_{t-1} + s_t G^t,\quad &\forall t > 0,\\
		U_0 \coloneqq 0,\quad U_t \coloneqq U_{t-1} + s_t G^t \circ
		G^t,\quad &\forall t > 0,
	\end{cases}
	\quad
	s_t \coloneqq \eta_t \sqrt{t}, 
\end{equation}
where $\circ$ denotes the Hadamard (pointwise) product in $\gE$.
We then construct the preconditioner $P^t$ and the weight sum
$\alpha_t$ by
\begin{equation}
\label{eq:ramda}
P^t \coloneqq \Diag(\sqrt[3]{U^t} + \epsilon),
		\quad \alpha_t
		\coloneqq \sum\nolimits_{k=1}^t s_k,
\end{equation}
where $\epsilon > 0$ is a (usually small) constant for numerical
stability and $\Diag(\cdot)$ is the diagonal matrix whose diagonal
entries are the elements of the input vector.
The update direction is then obtained by (approximately) solving the following subproblem.
\begin{equation}
		\hat W^{t} \approx \argmin_W\, \big(Q_t(W) \coloneqq
		\alpha_{t}\psi(W) + \inprod{V^t}{W}
		+ \frac{1}{2} \inprod{W - W^0}{P^t (W - W^0)}\big),
	\label{eq:prox}
\end{equation}
where $W^0$ is the initial point.
Details regarding \cref{eq:prox} and how to solve it
are deferred to \cref{sec:solver}.
The iterate is then updated by averaging $\hat W^t$ and $W^{t-1}$
with some $c_t \in [0,1]$:
\begin{align}
\label{eq:update}
	W^t  = \left( 1 - c_t \right) W^{t-1} + c_t \hat W^t.
\end{align}

The choice of $P^t$ in \cref{eq:ramda} that uses the accumulated square of
the stochastic gradient norm as the preconditioner is the key to
adaptivity and is widely seen in adaptive methods such as Adagrad
\cite{JD11a}, while the choice of the cubic root instead of the square
root is motivated by the impressive numerical performance of \madgrad
of \cite{DA21a} for smooth problems without a regularization term.
The averaging step in \cref{eq:update} with $c_t \neq 1$ can be interpreted as incorporating
a momentum term in the non-regularized non-adaptive case
\citep{tao2018primal,JS20a}.

\begin{algorithm}[tb]
\LinesNumbered
\DontPrintSemicolon
\caption{\ramda $(W^0, T, T_2, \epsilon, \{\eta_t\}, \{c_t\}, \{\epsilon_t\})$}
\label{alg:framework}

$V^0 \leftarrow 0,\quad U^0 \leftarrow 0, \quad \alpha_0 \leftarrow 0$

\For{$t=1,\dotsc,T$}
{
	Sample $\xi_t \sim \gD,
	\quad s_{t} \leftarrow \eta_t \sqrt{t},
	\quad \alpha_t \leftarrow \alpha_{t-1} + s_{t},\quad
	 G^t \leftarrow \nabla f_{\xi_t}(W^{t-1})$

	Compute $V^t, U^t$ by \cref{eq:da} and
	construct $P^t$ by \cref{eq:ramda}, and $\theta_t \gets
	\max(\diag(P^t))^{-1}$

	Compute $\hat W^t$ in \cref{eq:prox} by \pg$(W^{t-1}, W^0,
	\alpha_t^{-1} V^t,\alpha_t^{-1} P^t, \alpha_t\theta_t,
	 T_2,\epsilon_t)$

	 Update $W^t$ by \cref{eq:update}
}
\Output{$W^T$}
\end{algorithm}

\section{Subproblem Solver}
\label{sec:solver}
Given an iterate $W^{t-1}$, a momentum term $m_t$, a preconditioner $P^t$, and
a stepsize $\eta_t$,
existing regularized adaptive stochastic gradient algorithms for
\cref{eq:f} can be summarized in the following form
\citep{JY20a}:
\begin{equation}
	W^{t} = \argmin_W\, \big(\hat Q_t(W) \coloneqq \inprod{m_t}{W}
        + \frac{1}{2 \eta_t} \inprod{W - W^{t-1}}{P^t (W - W^{t-1})} 
        + \psi(W)\big),
	\label{eq:proxgen}
\end{equation}
whose form is similar to \cref{eq:prox}.
When the preconditioner $P^t$ is a multiple of the identity matrix
like in the case of \cite{ZSH21a}, the exact subproblem solution of
\cref{eq:prox} can be efficiently computed through the proximal
operator associated with the regularizer.
However, a major difficulty for realizing regularized adaptive methods,
including the proposed \ramda and the framework of \cite{JY20a}
whose preconditioners are not a multiple of the identity, is
that except for few special regularizers, the subproblem usually has no
closed-form solution.
We therefore consider using approximate solutions of the subproblem.

We propose to apply a few iterations of proximal gradient (PG)
\citep[see, \textit{e.g.},][]{BecT09a, Nes13a} to
approximately solve the subproblems in \cref{eq:prox,eq:proxgen} when no
closed-form solution is available, and we will show theoretically and
empirically in the following sections that the inexactness of such
approximate solutions has barely any effects on the theoretical
guarantees and the final model quality.
For the inexactness of the approximate solution in \cref{eq:prox}, we
require
\begin{equation}
	\min_{s \in \partial Q_t(\hat W^{t})} \norm{s} \leq \epsilon_t, \quad
	Q_t(\hat W^{t}) \leq Q_t(W^{t-1}),
	\label{eq:inexact}
\end{equation}
for some pre-specified $\epsilon_t$, where $\partial Q_t(W^{t+1})$ is
the (limiting) subdifferential \citep[see, \textit{e.g.},
][Definition~8.3]{RocW09a}.
This condition can be easily checked using information available in
the PG iterations.
For the sake of time efficiency, we also impose an upper limit
for the number of PG iterations.
Likewise, when applying our subproblem solver to \cref{eq:proxgen}, we
enforce \cref{eq:inexact} but with $Q_t$ replaced by $\hat Q_t$ and
$\hat W^t$ by $W^t$.
We focus on the case of diagonal and positive $P^t$, and thus the largest
eigenvalue $\max(\diag(P^t))$, where $\diag(\cdot)$ is
the vector formed by the diagonal entries of the input matrix, can be
calculated easily and used to compute a step size guaranteeing
sufficient objective decrease.
For cases in which this value is difficult to obtain, one can apply a
simple backtracking linesearch for the subproblem to find a suitable
step size efficiently.
This PG subproblem solver is summarized in \cref{alg:solver}.
To guarantee convergence for both our algorithm and the framework of
\cite{JY20a}, our analysis in \cref{sec:analysis} requires that
$\{\epsilon_t\}$ satisfy
\begin{equation}
	\bar \epsilon \coloneqq \sum\nolimits_{t=0}^\infty \epsilon_t^2 < \infty.
	\label{eq:eps}
\end{equation}
We will show in \cref{sec:analysis} that
\cref{eq:inexact} holds after at most $O(\epsilon_t^{-2})$
iterations of \cref{alg:solver}.

\begin{algorithm}[tb]
\DontPrintSemicolon
\caption{PG$(Z^0, W^0, V, P, \theta, T_2, \hat \epsilon)$}
\label{alg:solver}
\lIf{$\psi$ is nonconvex}{
$\theta \gets \theta / 2$}
\For{$j=1,\dotsc,T_2$} 
{
	$Z^j \leftarrow \prox_{\psi}(Z^{j-1} - \theta (V + P (Z^{j-1} -
	W^0)))$

	\lIf{\cref{eq:inexact} holds with $\epsilon_t = \hat \epsilon$ and
$\hat W^t = Z^j$}{$Z^{T_2} \gets Z^j$, and break}
}
\Output{$Z^{T_2}$}
\end{algorithm}

\section{Analysis}
\label{sec:analysis}

This section discusses theoretical guarantees for \ramda and the
proposed subproblem solver in \cref{alg:solver}.
We also prove convergence guarantees for applying \pg
to approximately solve \cref{eq:proxgen} for the framework of
\cite{JY20a}.
All proofs are in the appendices.
Some of our results are inspired by \cite{ZSH21a}, but with the added
inexactness in \cref{eq:prox} and the adaptiveness for the
preconditioner, the analysis is nontrivial.
Recall that we assume that $f_{\xi}$ is  differentiable only almost
everywhere but not everywhere,
which conforms with widely-used network structures like ReLU-type
activations.

We first show that \cref{eq:inexact} can be attained by
\cref{alg:solver}
and that the point of convergence of \ramda is almost surely a
stationary point. %

\begin{theorem}
	Assume that \cref{eq:prox} and \cref{eq:proxgen} has at least one
	optimal solution with a finite optimal objective value.
	Given $\epsilon_t > 0$, the number of iterations
	\cref{alg:solver} takes to satisfy \cref{eq:inexact} for
	both \cref{eq:prox} and \cref{eq:proxgen} is
	$O(\log(\epsilon_t^{-1}))$
	when $\psi$ is convex and $O(\epsilon_t^{-2})$ when $\psi$ is nonconvex.
	\label{thm:subproblem}
\end{theorem}

\begin{theorem}
\label{thm:stationary}
Consider $\{\hat W^t\}$ generated by \cref{alg:framework} for
\cref{eq:f}, with \cref{eq:inexact} and $\{c_t\}$ and $\{\epsilon_t\}$
satisfying $\sum c_t = \infty$ and \cref{eq:eps}.
Assume there is $L \geq 0$ such that for any $\xi$, $f_\xi$ is almost
surely $L$-Lipschitz-continuously-differentiable,
so the expectation is also $L$-Lipschitz-continuously-differentiable,
there is $C \geq 0$ such that $\E_{\xi_t \sim \gD}\norm{\nabla
	f_{\xi_t}\left( W^{t-1} \right)}^4 \leq C$ for all $t$,
and that the set of stationary points $\gZ \coloneqq \{W \mid 0 \in
\partial F(W) \}$ is nonempty.
For any given $W^0$, consider the event that $\{\hat W^t\}$ converges to a
point $\bar W$ (each event corresponds to a different $\bar W$).
If $\partial \psi$ is outer semicontinuous at $\bar W$,
this event has a nonzero probability, and $\{\eta_t\}$ satisfies
\begin{equation*}
	\sum s_t \alpha_t^{-1} = \infty,\quad \sum \left(s_t \alpha_t^{-1}\right)^2 < \infty,
	\norm{W^{t+1} - W^t}\left(s_t\alpha_t^{-1}\right)^{-1} \as 0,
\end{equation*}
then we have that $\bar W \in \gZ$ with probability one conditional on
this event.
Moreover, $\{W^t\}$ also converges to this stationary point $\bar W$.
\end{theorem}
Usually, convergence to a point requires some further regularity
conditions like the Kurdyka--{\L}ojasiewicz condition and boundedness
of the iterates.
However, existing frameworks regarding iterates convergence using such
conditions also require the method analyzed to have a
subgradient-descent-like behavior and to be a descent algorithm.
Neither of these hold true even for the basic stochastic gradient
algorithm, and we leave the analysis for this part as a challenging future work.

Our next key result shows that after a finite number of
iterations, iterates of \ramda all possess the same structure as that of the
point of convergence $\bar W$.
For this end, we first need to introduce the notions of partial
smoothness and prox-regularity,
and impose these assumptions on $\psi$ at $\bar W$.
\begin{definition}[Partial Smoothness \citep{Lew02a,HarL04a}]
	\label{def:ps}
	A function $\psi$ is partly smooth at a point $\bar W$ relative to a
	set $\gM_{\bar W} \ni \bar W$ if
	\begin{enumerate}[leftmargin=*,topsep=-16pt,itemsep=-7pt]
			\item Around $\bar W$, $\gM_{\bar W}$ is a $\mathcal{C}^2$-manifold and
				$\psi|_{\gM_{\bar W}}$ is $\mathcal{C}^2$.
			\item $\psi$ is regular
 (finite and the Fr\'{e}chet subdifferential coincides with the
 limiting Fr\'{e}chet subdifferential) at all points $W \in \gM_{\bar W}$
 near $\bar W$, with $\partial \psi(W) \neq \emptyset$.
		\item The affine span of $\partial \psi(\bar W)$ is a translate
			of the normal space to $\gM_{\bar W}$ at $\bar W$.
		\item $\partial \psi$ is continuous at $\bar W$ relative to $\gM_{\bar W}$.
	\end{enumerate}
\end{definition}
We often call $\gM_{\bar W}$ the active manifold at $\bar W$.
Locally, this manifold represents all points near $\bar W$ that share the same
structure induced by the regularized as $\bar W$.
Therefore, finding the active manifold is equivalent to finding the
locally optimal structure.
\begin{definition}[Prox-regularity \citep{RP96a}]
A function $\psi$ is prox-regular at $\bar W$ for
$V^* \in \partial \psi(\bar W)$ if $\psi$ is locally lower
semi-continuous around $\bar W$, finite at $\bar W$, and there is $\rho > 0$ such that
$\psi(W_1) \geq \psi(W_2) + \inprod{V}{W_1 - W_2} - \frac{\rho}{2}
\norm{W_1 - W_2}^2$ for every $W_1,W_2$ near $\bar W$ with
$\psi(W_2)$ close to $\psi(\bar W)$ and $V \in \partial \psi(W_2)$ close
to $V^*$. $\psi$ is prox-regular at $\bar W$ if it is prox-regular for
all $V \in \partial \psi(\bar W)$.
\end{definition}
\begin{theorem}
\label{thm:identify}
Consider \cref{alg:framework} with the conditions in \cref{thm:stationary}
satisfied.
Consider the event of $\{\hat W^t\}$ converging to a certain point $\bar W$
as in \cref{thm:stationary}.
If the probability of this event is nonzero; $\psi$ is prox-regular
and subdifferentially continuous at $\bar W$ and partly smooth at
$\bar W$ relative to the active $\mathcal{C}^2$ manifold $\gM_{\bar
W}$; $\partial \psi$ is outer semicontinuous at $\bar W$; and the
nondegeneracy condition
	$-\nabla f\left( \bar W \right) \in \textup{relative interior of}\; \partial \psi
	\left( \bar W \right)$
holds at $\bar W$,
then conditional on this event, almost surely there is $T_0 \geq 0$
such that
\begin{equation*}
	\hat W^t \in \gM_{\bar W},\quad \forall t \geq T_0.
\end{equation*}
\end{theorem}
We note particularly that convex and weakly-convex \citep{EAN73a} functions are all
regular, prox-regular, and subdifferentially continuous everywhere.

We also show that our subproblem solver and condition can be
effectively applied to the framework of \cite{JY20a} while retaining
the same convergence guarantees.
As mentioned in \cref{sec:related}, our result is much stronger
than that of \cite{deleu2021structured} for having no unrealistic
smoothness requirement on $\psi$ and using an implementable inexactness
condition.

\begin{theorem}
	For the framework in \cite{JY20a} with the subproblem solved
	approximately by \cref{alg:solver} such that \cref{eq:inexact}
	holds with $\{\epsilon_t\}$ satisfying \cref{eq:eps}.
	Then Theorem~1 of \cite{JY20a} still holds, but with the constants
	$\{Q_i\}$ being also dependent on $\bar \epsilon$.
	\label{thm:proxgen}
\end{theorem}

\section{Experiments}
\label{sec:exp}
This section examines the practical performance of \ramda for training
structured neural networks.
As sparsity is arguably one of the most widely adopted structures in
machine learning, we follow \cite{wen2016learning} to consider
structured sparsity as the representative structure in our experiments.
Particularly, we employ the group LASSO regularization \citep{yuan2006model} to encourage group sparsity and disable weight decay in all experiments, except for the dense baselines.
We begin from examining the efficiency and effectiveness of
\pg for both \ramda and existing regularized adaptive
methods.
We then consider tasks in computer vision, language modeling, and
speech to compare the following algorithms using Pytorch.
\begin{itemize}[leftmargin=*]
\parskip 0pt
\topsep 0pt
\partopsep 0pt
\itemsep 0pt
\item \ramda: The proposed \cref{alg:framework}.
\item \rmda \citep{ZSH21a}%
\item \proxsgd \citep{yang2019proxsgd}%
\item \proxgen \citep{JY20a}:  %
	We follow their experiments to use
	\adamw and apply our \pg as
	the subproblem solver.
\item \proxssi \citep{deleu2021structured}
\end{itemize}
These algorithms are introduced in \cref{sec:related} and also further summarized in \cref{sec:sumalg}.
For each task, we also provide for reference a baseline that does not include a
group LASSO regularizer in the training (SGD with momentum (\msgd) for
computer vision, and \adam for the other two), but our
comparison is only among those for training structured models.
Our code for reproducing the experiments and the hyperparameter settings are available
at \url{https://github.com/ismoptgroup/ramda_exp/}.
Additional details of the stability of the structure
(level of structured sparsity here) over epochs of \ramda
are available in \cref{app:plot}.

We use two criteria for comparison:
1. Model predictive ability, and 2. Structured sparsity level.
The former is task-dependent and thus specified in each experiment.
Regarding the latter,
sparsifying neural networks while preserving its performance
requires prior knowledge of model design. A common approach is
retaining certain parameters during the training process, and
we adhere to this convention such that the bias, batch
normalization \citep{ioffe2015batch}, layer normalization
\citep{JLB16a}, and embedding layers do not have any sparsity-inducing
regularization imposed on them \citep{deleu2021structured, AP21a}.
For the rest,
we adopt channel-wise grouping for
convolutional layers, input-wise grouping for fully-connected
and LSTM layers during the training phase.
For evaluation, our structured sparsity is calculated using the weighted group sparsity
with the weights proportional to the number of parameters in each group.

We run each experiment with three different random
initializations and show the mean and standard deviation of the
validation predictive performance and the structured sparsity of the
final model of all methods.

\paragraph{Subproblem}
\begin{table}[tb]
\caption{Weighted group sparsity and validation accuracy of
	different subproblem stopping criteria.}
\label{tbl:subproblem}
\begin{center}
\begin{tabular}{@{}l|l|rr|rr@{}}
& & \multicolumn{2}{c|}{No early stopping} & \multicolumn{2}{c}{Early stopping}\\
\hline
Model/Data & Algorithm & Accuracy & Sparsity & Accuracy & Sparsity \\
\hline
VGG19 / &
\proxgen & 92.7 $\pm$ 0.2\% & 88.8 $\pm$ 0.0\% & 92.7 $\pm$ 0.1\% & 86.9 $\pm$ 0.4\% \\
CIFAR10 &
\ramda & 92.7 $\pm$ 0.2\% & 86.7 $\pm$ 0.3\% & 92.9 $\pm$ 0.2\% & 86.3 $\pm$ 0.4\% \\
\hline
ResNet50 / & \proxgen & 73.6 $\pm$ 0.1\% & 74.7 $\pm$ 0.6\% & 74.0 $\pm$ 0.1\% & 67.6 $\pm$ 3.1\% \\
CIFAR100 & \ramda & 69.9 $\pm$ 1.5\% & 69.5 $\pm$ 2.1\% & 71.2 $\pm$ 1.4\% & 67.5 $\pm$ 1.6\%
\end{tabular}
\end{center}
\end{table}

We start from showing the effectiveness of our proposed subproblem
solver for \ramda and \proxgen.
For both approaches, we use Theorem~2 of \cite{deleu2021structured} to
safely screen out a portion of groups that will be zero at the optimal
subproblem solution,
and opt for the \pg algorithm to solve the remaining parts.
We consider two practical stopping criteria for \pg: 1. Running until
it reaches the maximum iterations (no early stopping), and 2.
Terminate when the subproblem objective improvement is small
(early stopping).
For the former, we set the maximum to $100$.
For the latter, we terminate \pg early if $(Q_t(Z^{j-1}) - Q_t(Z^{j})) /
(|Q_t(Z^{j}| + 1) < 10^{-8}$ is reached.
Moreover, to ensure incorporation of the preconditioner into
\proxgen, we set its minimum PG iterations to $2$.
We examine how these stopping criteria affect the final model of
\ramda and \proxgen using image classification problems of a smaller
scale.
From \cref{tbl:subproblem}, we see that early stopping does not affect
the outcome much.
Given that early stopping is more efficient, we will adopt it in all subsequent experiments.

Next, we compare \proxgen with \proxssi (these two only differ in the
subproblem solver) to examine the efficiency and performance
differences between solving the subproblems approximately and (almost) exactly
in \cref{tbl:proxssi}.
We see that our solver is around 3X faster than
\proxssi, and the model qualities are similar.
We thus exclude \proxssi from our
comparisons in the following experiments due to its excessively
lengthy running time, especially for large-scale tasks.

\begin{table}[tb]
\caption{Weighted group sparsity, validation accuracy and time/epoch of 
        \proxssi and \proxgen for CIFAR10/CIFAR100. 
        We report the average time/epoch using one NVIDIA V100 GPU.}
\label{tbl:proxssi}
\begin{center}
\begin{tabular}{@{}lrrr|rrr@{}}
Algorithm & Accuracy & Sparsity & Time & Accuracy & Sparsity & Time \\
\hline
\multicolumn{3}{c}{VGG19/CIFAR10} & \multicolumn{3}{c}{VGG19/CIFAR100} \\ 
\hline
\proxssi & 92.8 $\pm$ 0.1\% & 88.4 $\pm$ 0.2\% & 79s & 67.3 $\pm$ 0.1\% & 78.6 $\pm$ 0.3\% & 79s \\
\proxgen & 92.8 $\pm$ 0.0\% & 86.6 $\pm$ 0.1\% & 24s & 68.1 $\pm$ 0.4\% & 75.5 $\pm$ 0.2\% & 26s \\
\hline
\multicolumn{3}{c}{ResNet50/CIFAR10} & \multicolumn{3}{c}{ResNet50/CIFAR100} \\
\hline
\proxssi & 94.0 $\pm$ 0.1\% & 83.7 $\pm$ 0.6\% & 260s & 73.7 $\pm$ 0.4\% & 70.4 $\pm$ 0.7\% & 251s \\
\proxgen & 94.1 $\pm$ 0.1\% & 80.4 $\pm$ 0.4\% & 70s & 73.6 $\pm$ 0.4\% & 65.5 $\pm$ 3.6\% & 74s \\
\hline
\end{tabular}
\end{center}
\end{table}

\paragraph{Image Classification}

We conduct a classical computer vision task of training ResNet50
\citep{he2016deep} with the ILSVRC 2012 ImageNet dataset \citep{OR15a}.
The result in \cref{tbl:cv} shows
that \ramda attains the best validation accuracy and
structured sparsity simultaneously.

\begin{table}[tb]
\caption{Weighted group sparsity and validation accuracy on
ImageNet/ResNet50.}
\label{tbl:cv}
\renewrobustcmd{\bfseries}{\fontseries{b}\selectfont}
\renewrobustcmd{\boldmath}{}
\newrobustcmd{\B}{\bfseries}
\begin{center}
\sisetup{detect-weight,mode=text}
\begin{tabular}{@{}lrr@{}}
Algorithm & Accuracy & Sparsity \\
\hline
\msgd & 77.14 $\pm$ 0.04\% & \multicolumn{1}{c}{-} \\
\hline
\hline
\ramda & \B{74.53 $\pm$ 0.10\%} & \B{29.19 $\pm$ 0.94\%}\\
\proxsgd & 73.50 $\pm$ 0.20\% & 17.54 $\pm$ 1.26\% \\
\proxgen & 74.17 $\pm$ 0.08\% & 20.29 $\pm$ 0.22\% \\
\rmda & 74.47 $\pm$ 0.08\% & 25.20 $\pm$ 1.69\%
\end{tabular}
\end{center}
\end{table}

\paragraph{Language Modeling}

For language modeling, we train Transformer-XL (base) \citep{ZD19a} using the
WikiText-103 dataset \citep{SM17a}.
Transformer-XL is comprised of embedding and non-embedding layers, and
in the PyTorch implementation, the non-embedding layers are built
using linear and layer-normalization layers.
We apply group LASSO regularization to the linear layers, and
present in \cref{tbl:Transformer-XL} the perplexity and the weighted
group sparsity of the models trained.
We see that \ramda gives the best perplexity and structured
sparsity simultaneously.

\begin{table}[tb]
\caption{Weighted group sparsity and validation perplexity on
Transformer-XL with WikiText-103.}
\label{tbl:Transformer-XL}
\begin{center}
\renewrobustcmd{\bfseries}{\fontseries{b}\selectfont}
\renewrobustcmd{\boldmath}{}
\newrobustcmd{\B}{\bfseries}
\begin{center}
\begin{tabular}{@{}lrrr@{}}
Alg. & Perplexity & Sparsity & Time/epoch \\
\hline
\adam & 23.00 $\pm$ 0.05 & \multicolumn{1}{c}{-} & 6261 $\pm$ 21s\\
\hline
\hline
\ramda & \B{26.97 $\pm$ 0.10} & \B{36.2 $\pm$ 0.3\%} & 6954 $\pm$ 30s\\
\proxsgd & 27.42 $\pm$ 0.02 & 33.1 $\pm$ 1.5\% & 6167 $\pm$ 12s\\
\proxgen & 27.49 $\pm$ 0.19 & 30.5 $\pm$ 0.6\% & 6652 $\pm$ 21s\\
\rmda & 27.10 $\pm$ 0.08 & 36.0 $\pm$ 2.7\% & 6184 $\pm$ 20s
\end{tabular}
\end{center}
\end{center}
\end{table}
\paragraph{Speech Synthesis}

We consider Tacotron2 \citep{JS18a} for speech synthesis 
on the LJSpeech dataset \citep{KI17a}. 
We apply regularization to the convolutional, LSTM, and linear layers
of Tacotron2 and show the results in \cref{tbl:speech}.
Clearly, \ramda gives the lowest validation loss and the highest
group sparsity.

\begin{table}[tb]
\caption{Weighted group sparsity and validation loss on
Tacotron2 with LJSpeech.}
\label{tbl:speech}
\begin{center}
\renewrobustcmd{\bfseries}{\fontseries{b}\selectfont}
\renewrobustcmd{\boldmath}{}
\newrobustcmd{\B}{\bfseries}
\begin{tabular}{@{}lrrrr@{}}
Alg. & Loss & Sparsity & Time/epoch \\
\hline
\adam & 0.39  $\pm$  0.02 & \multicolumn{1}{c}{-} & 431  $\pm$  2s \\
\hline
\hline
\ramda & \textbf{0.44  $\pm$  0.01} & \textbf{52.9  $\pm$  1.6\%} & 443  $\pm$  1s \\
\proxsgd & 0.50  $\pm$  0.00 & 34.3  $\pm$  1.6\% & 431  $\pm$  0s \\
\proxgen & 0.45  $\pm$  0.01 & 45.6  $\pm$  0.9\% & 438  $\pm$  2s \\
\rmda & 0.46  $\pm$  0.01 & 45.9  $\pm$  1.7\% & 431  $\pm$  2s
\end{tabular}
\end{center}
\end{table}

\paragraph{Time Efficiency}
In \cref{tbl:Transformer-XL,tbl:speech}, we see that although \ramda
and \proxgen have more difficult subproblems without a closed-form
solution to solve, our proposed \pg solver is highly efficient such
that the running time of them is still close to other approaches,
making these regularized adaptive approaches practically feasible.

\paragraph{Summary}
In summary, thanks to its adaptive nature (for better predictive
performance) and its ability of manifold identification (for higher
structured sparsity),
\ramda is superior to state of the art on modern language
modeling and speech synthesis tasks as well as the ImageNet problem.
We also observe from the plots in the appendices that it is possible to
further improve the sparsity level of \ramda if we run it for more
epochs.

\section{Conclusions}
\label{sec:conclusions}
In this work, we proposed a regularized dual averaging method with
adaptiveness, \ramda, for training structured neural networks.
Our method outperforms state of the art on modern architectures
including LSTM and transformers as well as the ImageNet problem.
We also proposed a subroutine with strong convergence guarantees to
approximately solve the regularized subproblem of both our method and
an existing framework efficiently.
Extensive experiments on group sparsity showed that our subproblem
solver can greatly reduce the training time for existing methods, and
our proposed \ramda achieves simultaneously higher structured sparsity
ratio and better prediction performance than existing methods.
Implementation of our method is available at
\url{https://www.github.com/zhisyuan1214/RAMDA/}.

\section*{Acknowledgement}
Ching-pei's research is supported in part by the JSPS Grant-in-Aid
for Research Activity Start-up 23K19981 and Grant-in-Aid
for Early-Career Scientists 24K20845.
\bibliographystyle{plainnat}
\bibliography{ramda}

\appendix
\part{Appendices} %
\parttoc %
\section{More Experiment Details}
\label{sec:sumalg}
This section describes details of our implementation of \ramda and
the setting of the experiments conducted in \cref{sec:exp}.

\subsection{Implementation and Hyperparameter Setting of \ramda}
Similar to \cite{ZSH21a}, we introduce a restart strategy to the
implementation of \ramda.
During each stage of the training, the learning rate $\eta_t$ and the momentum
factor $c_t$ are fixed.
Once the epoch count enters the next stage, we reset the counter $t$
to $1$ and use the output parameter $W^T$ from the previous round as
the new input parameter $W^0$ to the same algorithm,
set $\alpha_t, V^t$ and $U^t$ to $0$, but keep the scheduling for
$\eta$ and $c$ going without resetting them,
and decrease $\epsilon$ by a factor.
We initialize $c_t$ as either $0.1$ or $0.01$,
depending on the problems, and
use a constant $c_t$ until the final stage, where we
gradually increase it by
\begin{equation*}
		c_t = \min(c\sqrt{i}, 1),
\end{equation*}
where $i$ counts the training steps executed at this final stage.
This momentum strategy is applied to both \ramda and
\rmda in our experiments.

\subsection{Details of the Algorithms Compared}
In \cref{tbl:algorithm}, we summarize details of the algorithms
compared in \cref{sec:exp}.

\begin{table*}[tbh]
\centering
\caption{Algorithms used in the experiments.}
\label{tbl:algorithm}
\begin{tabular}{l|l|l}
    Algorithm & Unregularized counterpart & Subproblem \\\hline 
        \ramda & \madgrad \citep{DA21a} & \pg \\\hline 
    \rmda & \mda \citep{JS20a} & Closed-form solution \\\hline 
    \proxsgd & \msgd & Closed-form solution \\\hline 
    \proxgen & \adamw \citep{IL19a} & \pg \\\hline 
        \proxssi & \adamw \citep{IL19a} & Newton-Raphson
\end{tabular}
\end{table*}

\subsection{Computational Resource}
\label{sec:gpu}
We conduct all experiments utilizing NVIDIA TESLA V100 (32 GB) GPUs.
We employ eight V100 GPUs for each run of the ImageNet experiments.
For all other experiments, we utilize a single V100 GPU per run.

\section{Proofs}
\label{app:proof}
This section provides proofs of the theoretical results stated in
\cref{sec:analysis}.
We restate these results and provide their corresponding proofs right
after each statement.

\begin{reptheorem}{thm:subproblem}
	Assume that \cref{eq:prox} and \cref{eq:proxgen} has at least one
	optimal solution with a finite optimal objective value.
	Given $\epsilon_t > 0$, the number of iterations of
	\cref{alg:solver} takes to satisfy \cref{eq:inexact} for
	both \cref{eq:prox} and \cref{eq:proxgen} is $O(\epsilon_t^{-1})$
	when $\psi$ is convex and $O(\epsilon_t^{-2})$ when $\psi$ is nonconvex.
\end{reptheorem}
\begin{proof}
We use the notation
\[
	\bar Q_t(Z) = f_t(Z) + \psi(Z)
\]
to unify the two objective function $Q_t / \alpha_t$ and $\hat Q_t$,
where $f_t$ is the smooth part and we define the Lipschitz constant of
$\nabla f_t$ as $L$.

At each iteration, \pg solves the following subproblem
\[
	Z^{j+1} \in \argmin_{Z}\quad \inprod{\nabla f_t(Z^j)}{Z - Z^j} +
	\frac{1}{2 \theta_t}\norm{Z - Z^j}^2 + \psi(Z),
\]
and thus from the first-order optimality conditions, clearly we have
\[
	\nabla f_t(Z^{j+1}) - \nabla f_t(Z^{j}) - \frac{1}{\theta_t}\left(
	Z^{j+1} - Z^j\right) \in \partial \bar Q_t(Z^{j+1}).
\]
We thus have from the Lipschitz continuity of $\nabla f_t$ that
\begin{equation}
	\label{eq:subgradbound}
	\min_{s \in \partial \bar Q_t(Z^{j+1})} \norm{s} \leq \norm{
		\nabla f_t(Z^{j+1}) - \nabla f_t(Z^{j})} +
		\frac{1}{\theta_t}\norm{
		Z^{j+1} - Z^j}
		\leq \left( L + \theta_t^{-1} \right) \norm{Z^{j+1} - Z^j}.
\end{equation}
Note that $\bar Q_t$ is lower bounded, say by $\bar Q_t^*$, and has at
least one solution $Z^*$ (unique when $\psi$ is convex).

In the case that $\psi$ is convex, we know that $\theta_t = 1/L$, and
\cref{eq:ramda} clearly shows that the subproblem objective $\bar Q$ is
$\epsilon$-strongly convex.
Therefore, standard analysis of proximal gradient
\citep[see, for example,][Lemma~10.4 and iTheorem~10.29]{Bec17a} gives that
\begin{gather}
	\label{eq:norm1}
	\frac{L}{2} \norm{Z^{j+1} - Z^j} \leq \bar Q_t(Z^j) - \bar
	Q_t(Z^{j+1}),\quad \forall j \geq 0,\\
	\label{eq:norm2}
	\bar Q_t(Z^j) - \bar Q_t(Z^*) \leq \frac{L}{2}\left(1 -
	\frac{\epsilon}{L}\right)^j \norm{Z^0 - Z^*}^2, \quad \forall j
	\geq 1.
\end{gather}
The combination of \cref{eq:subgradbound,eq:norm1,eq:norm2} shows that
it takes $O(\log \epsilon_t^{-1})$ iterations for \pg to reach the
required precision of $\epsilon_t$.

When $\psi$ is nonconvex, we have that $\theta_t = 1/ (2L)$ and
standard analysis \citep[Section~5.1]{AttBS13a} gives
\begin{equation}
	\min_{k = 0,1,\dots,j} \norm{Z^{j+1} - Z^j} \leq
	\frac{C}{\sqrt{j}}
	\label{eq:norm3}
\end{equation}
for some constant $C$ depending on $L$ and $\bar Q_t(W^t) - \bar
Q_t^*$.
Therefore, \cref{eq:norm3} and \cref{eq:subgradbound} show that it
takes $O(\epsilon_t^{-2})$ iterations to reach the desired precision.
\end{proof}

\begin{reptheorem}{thm:stationary}
Consider $\{\hat W^t\}$ generated by \cref{alg:framework} for
\cref{eq:f}, with \cref{eq:inexact} and $\{c_t\}$ and $\{\epsilon_t\}$
satisfying $\sum c_t = \infty$ and \cref{eq:eps}.
Assume there is $L \geq 0$ such that for any $\xi$, $f_\xi$ is almost
surely $L$-Lipschitz-continuously-differentiable,
so the expectation is also $L$-Lipschitz-continuously-differentiable,
there is $C \geq 0$ such that $\E_{\xi_t \sim \gD}\norm{\nabla
	f_{\xi_t}\left( W^{t-1} \right)}^4 \leq C$ for all $t$,
and that the set of stationary points $\gZ \coloneqq \{W \mid 0 \in
\partial F(W) \}$ is nonempty.
For any given $W^0$, consider the event that $\{\hat W^t\}$ converges to a
point $\bar W$ (each event corresponds to a different $\bar W$).
If $\partial \psi$ is outer semicontinuous at $\bar W$,
this event has a nonzero probability, and $\{\eta_t\}$ satisfies
\begin{align}
	& \sum s_t \alpha_t^{-1} = \infty,\quad \sum \left(s_t \alpha_t^{-1}\right)^2 < \infty, \\
	& \norm{W^{t+1} - W^t}\left(s_t\alpha_t^{-1}\right)^{-1} \as 0,
\label{eq:schedule}
\end{align}
then we have that $\bar W \in \gZ$ with probability one conditional on
this event.
Moreover, $\{W^t\}$ also converges to this stationary point $\bar W$.
\end{reptheorem}
\begin{proof}
First, we prove that when $\{\hat W^t\}$ converges to $\bar W$,
$\{W^t\}$ also converges to $\bar W$.
From \cref{eq:update}, we have that
\begin{equation}
	\norm{W^t - \bar W} \leq (1 - c_t) \norm{W^{t-1} - \bar W} + c_t
	\norm{\hat W^t - \bar W}.
	\label{eq:normchange}
\end{equation}
Since $\hat W^t \rightarrow \bar W$, for any $\epsilon > 0$ we can
find an integer $t_\epsilon \geq 0$ such that $\|{\hat W^t - \bar
W}\| \leq \epsilon$ for all $t \geq t_{\epsilon}$.
Therefore, by deducting $\epsilon$ from both sides of
\cref{eq:normchange}, we get
\begin{equation*}
	\norm{W^t - \bar W} - \epsilon
	\leq \left(\prod_{k=t_{\epsilon}}^{t}(1 -
	c_t)\right) \left(\norm{W^{t_{\epsilon}-1}
- \bar W} - \epsilon\right)
	\leq \exp\left(-\sum_{k=t_{\epsilon}}^{t}
	c_t\right) \left(\norm{W^{t_{\epsilon}-1}
- \bar W} - \epsilon\right)
,\quad \forall t \geq t_{\epsilon}.
\end{equation*}
By letting $t$ approach infinity and noting that $\sum c_t = \infty$, we see that
\[
	\lim_{t \rightarrow \infty} \norm{W^t - \bar W} \leq \epsilon.
\]
Because $\epsilon$ is arbitrary, we see that $\|{W^t - \bar W}\|
\rightarrow 0$, and hence $\{W^t\}$ converges to $\bar W$.

Next, consider the update of $\alpha_t^{-1} U^t$, we can see from
\cref{eq:da} that
\begin{equation}
	\frac{U^t}{\alpha_t} = \frac{\alpha_{t-1}}{\alpha_t}
	\frac{U^{t-1}}{\alpha_{t-1}} + \frac{s_t
		\nabla f_{\xi_t}(W^{t-1})}{\alpha_t} = \left(1 -
		\frac{s_t}{\alpha_t}\right)\frac{U^{t-1}}{\alpha_{t-1}} +
		\frac{s_t}{\alpha_t}\nabla f_{\xi_t}(W^{t-1}).
		\label{eq:Uupdate}
\end{equation}
Moreover, the assumptions on $\eta_t$ satisfies all the required
conditions of Lemma~1 of \cite{Rus80a}.
We therefore apply Lemma~1  of \cite{Rus80a} to conclude that
\begin{equation}
	\frac{U^t}{\alpha_t} \xrightarrow{\text{a.s.}}
	\E_{\xi\sim \gD} \left[\nabla f_{\xi} \left( W^{t} \right) \circ
	\nabla f_{\xi} \left( W^{t} \right) \right].
	\label{eq:convU}
\end{equation}
The update for $\alpha_t^{-1} V^t$ has a form analogous to
\cref{eq:Uupdate}, and we have from Jensen's inequality that
\[
\E_{\xi_t \sim \gD}\norm{\nabla f_{\xi_t}\left( W^{t-1} \right)}^2
\leq \sqrt{\E_{\xi_t \sim \gD}\norm{\nabla f_{\xi_t}\left( W^{t-1}
\right)}^4} \leq \sqrt{C},
\]
implying that the second moment is also bounded in expectation.
We can therefore also apply Lemma~1  of \cite{Rus80a} to
$\alpha_t^{-1}V^t$ and conclude that
	\begin{equation}
		\frac{V^t}{\alpha_t} \xrightarrow{\text{a.s.}} \nabla \E_{\xi\sim \gD} \left[
		f_{\xi} \left( W^{t} \right) \right].
		\label{eq:convV}
	\end{equation}
We further notice that the union of two events that happens almost
surely is still an event that happens almost surely.

From \cref{eq:prox,eq:inexact}, we can see that there is a
sequence $\{z_t\}$ such that
	\begin{equation}
		\label{eq:firstorder}
		-\left(\frac{V^t}{\alpha_t} + \frac{z_t}{\alpha_t} +
		\frac{P^t}{\alpha_t}
		(\hat W^t - W^0)\right) \in \partial \psi(\hat W^t),\quad
		\norm{z_t} \leq \epsilon_t.
	\end{equation}
	Our assumption in \cref{eq:schedule} implies that
	$\alpha_t \rightarrow \infty$, which together with \cref{eq:eps}
	leads to
\begin{equation}
	\frac{z_t}{\alpha_t} \rightarrow 0.
	\label{eq:term2}
\end{equation}
From \cref{eq:convV}, that $\nabla \E_{\xi\sim \gD} \left[ f_{\xi}
\left( W \right) \right]$ is Lipschitz continuous, and that $ W^t
\rightarrow \bar W$ (which we have proven in the first part), we see that
\begin{equation}
	\frac{V^t}{\alpha_t} \xrightarrow{\text{a.s.}} \nabla  \E_{\xi\sim \gD} \left[
	f_{\xi}(\bar W)\right].
	\label{eq:term1}
\end{equation}
For the third term, we have from \cref{eq:convU,eq:ramda} that
\[
	\frac{P^t}{\alpha_t} = \alpha_t^{-\frac{2}{3}} \Diag\left(\sqrt[3]{\frac{U^t}{\alpha_t}}\right) +
	\frac{\epsilon}{\alpha_t} I.
\]
Again since $\alpha_t \rightarrow \infty$, the second term of the
equation above converges to $0$.
Therefore, by \cref{eq:convU}, we obtain that
\[
	\frac{P^t}{\alpha_t} \xrightarrow{\text{a.s.}}
	\alpha_t^{-\frac{2}{3}} \Diag\left(\sqrt[3]{ \E_{\xi\sim \gD}
	\left[\nabla f_{\xi} \left( W^{t} \right) \circ \nabla f_{\xi}
	\left( W^{t} \right) \right]}\right).
\]
Again from the continuity of $\nabla \E_{\xi\sim \gD} \left[ f_{\xi}
\left( W \right) \right]$ and that $\alpha_t \rightarrow \infty$,
we conclude that
\begin{equation}
	\label{eq:term3}
	\frac{P^t}{\alpha_t} \xrightarrow{\text{a.s.}}
	\alpha_t^{-\frac{2}{3}} \Diag\left(\sqrt[3]{ \E_{\xi\sim \gD}
	\left[\nabla f_{\xi} \left( \bar W \right) \circ \nabla f_{\xi}
	\left( \bar W \right) \right]}\right)
	\xrightarrow{\text{a.s.}} 0.
\end{equation}
Finally, using the outer semicontinuity of $\partial \psi(W)$ at $\bar W$, we
conclude from \cref{eq:firstorder,eq:term2,eq:term1,eq:term3} that
\begin{equation*}
0 \in \nabla \E_{\xi\sim \gD} \left[ f_{\xi} \left( \bar W\right)
\right] + \lim_{t \rightarrow \infty} \psi(\hat W^t)
\subseteq
\nabla \E_{\xi\sim \gD} \left[ f_{\xi} \left( \bar W\right)
\right] + \psi(\bar W) = \partial F(\bar W)
\end{equation*}
with probability one, showing that $\bar W$ is a stationary point almost
surely.
\end{proof}

\begin{reptheorem}{thm:identify}
Consider \cref{alg:framework} with the conditions in \cref{thm:stationary}
satisfied.
Consider the event of $\{\hat W^t\}$ converging to a certain point $\bar W$
as in \cref{thm:stationary}.
If the probability of this event is nonzero; $\psi$ is prox-regular
and subdifferentially continuous at $\bar W$ and partly smooth at
$\bar W$ relative to the active $\mathcal{C}^2$ manifold $\gM_{\bar
W}$; $\partial \psi$ is outer semicontinuous at $\bar W$; and the
nondegeneracy condition
\begin{equation}
	-\nabla f\left( \bar W \right) \in \textup{relative interior of}\; \partial \psi
	\left( \bar W \right)
	\label{eq:nod}
\end{equation}
holds at $\bar W$,
then conditional on this event, almost surely there is $T_0 \geq 0$
such that
\begin{equation}
	\hat W^t \in \gM_{\bar W},\quad \forall t \geq T_0.
	\label{eq:identify}
\end{equation}
In other words, the active manifold at $\bar W$ is identified by the
iterates of \cref{alg:framework} after a finite number of iterations almost
surely.
\end{reptheorem}
\begin{proof}
From \cref{eq:firstorder}, there exists a sequence $\{Y^t\}$ such
that
\begin{equation}
	\label{eq:Y}
	Y^t \in \partial \psi(\hat W^t),\quad
	\frac{V^t}{\alpha_t} + \frac{z_t}{\alpha_t} +
	\frac{P^t}{\alpha_t}
	(\hat W^t - W^0) + Y^t = 0, \quad \forall t.
\end{equation}
For notational ease, we denote
\begin{equation}
	\label{eq:notation}
	f(W) \coloneqq \E_{\xi\sim \gD} \left[ f_{\xi}(W)\right].
\end{equation}
From \cref{eq:Y}, we then get
\begin{equation}
	\nabla f(\hat W^t) - \frac{V^t}{\alpha_t} - \frac{z_t}{\alpha_t} -
	\frac{P^t}{\alpha_t}
	(\hat W^t - W^0) \in \partial F(\hat W^t).
	\label{eq:tobound}
\end{equation}
We aim to show that \[
	\dist(0, \partial F(\hat W^t)) \coloneqq \min_{Y \in \partial
	F(\hat W^t)} \norm{Y}
\]
converges to $0$ almost surely.
From \cref{eq:tobound}, we have
\begin{align}
	\nonumber
\dist(0, \partial F(\hat W^t))
&\leq \norm{ \nabla f(\hat W^t) - \frac{V^t}{\alpha_t} -
\frac{z_t}{\alpha_t} - \frac{P^t}{\alpha_t} (\hat W^t - W^0)} \\
	\nonumber
&\leq \norm{ \nabla f(\hat W^t) - \frac{V^t}{\alpha_t}} +
\norm{\frac{z_t}{\alpha_t}} + \norm{\frac{P^t}{\alpha_t} (\hat W^t -
W^0)}\\
&\leq \norm{ \nabla f(\hat W^t) - \frac{V^t}{\alpha_t}} +
\frac{\epsilon_t}{\alpha_t} + \norm{\frac{P^t}{\alpha_t} (\hat W^t -
W^0)},
	\label{eq:tobound2}
\end{align}
where we get the first inequality from the triangle inequality and the
second from \cref{eq:firstorder}.
According to \cref{eq:convV,eq:term3}, there are $\{A_t\}$ and $\{B_t\}$ such
that
\begin{equation}
	\label{eq:as}
	\begin{cases}
		\frac{V^t}{\alpha_t} &= \nabla f(W^t) + A_t, \quad \norm{A_t} \as 0\\
		\frac{P^t}{\alpha_t} &= \alpha_t^{-\frac23}
		\Diag\left( \sqrt[3]{\nabla f(W^t) \circ \nabla f(W^t)}
		\right) + B_t, \quad \norm{B_t} \as 0.
	\end{cases}
\end{equation}
Substituting the above two equations back to \cref{eq:tobound2}, we
obtain
\begin{align}
\nonumber
&~\dist(0, \partial F(\hat W^t))\\
\nonumber
\leq &~\norm{ \nabla f(\hat W^t) - \nabla f(W^t)} + \norm{A_t} +
\frac{\epsilon_t}{\alpha_t} + \left(
		 \alpha_t^{-\frac23}
		 \norm{\sqrt[3]{\nabla f(W^t) \circ \nabla f(W^t)}}_{\infty}
		 + \norm{B_t} \right)\norm{\hat W^t - W^0}\\
\leq &~L \norm{\hat W^t - W^t} + \norm{A_t} +
\frac{\epsilon_t}{\alpha_t} + \left(
		 \alpha_t^{-\frac23}
		 \norm{\sqrt[3]{\nabla f(W^t) \circ \nabla f(W^t)}}_{\infty}
		 + \norm{B_t} \right)\norm{\hat W^t - W^0}.
	\label{eq:tobound3}
\end{align}
From \cref{thm:stationary}, we know that $\hat W^t$ and $W^t$ both
converge to $\bar W$, so
\[
	\norm{\hat W^t - W^t} \leq \norm{\hat W^t - \bar W} + \norm{W^t -
	\bar W} \rightarrow 0.
\]
From \cref{eq:schedule,eq:eps}, we know that $\epsilon_t/\alpha_t
\rightarrow 0$.
Because $\hat W^t \rightarrow \bar W$, we also have that
\[
\norm{\hat W^t - W^0} \rightarrow \norm{\bar W - W^0} < \infty.
\]
From $W^t \rightarrow \bar W$, we have that
\[
	\norm{\sqrt[3]{\nabla f(W^t) \circ \nabla f(W^t)}}_{\infty}
	\rightarrow \norm{\sqrt[3]{\nabla f(\bar W) \circ \nabla f(\bar
	W)}}_{\infty} <
	\infty.
\]
Combining these results with \cref{eq:tobound3}, we conclude that
\begin{align*}
L \norm{\hat W^t - W^t} + \norm{A_t} +
\frac{\epsilon_t}{\alpha_t} + \left(
		 \alpha_t^{-\frac23}
		 \norm{\sqrt[3]{\nabla f(W^t) \circ \nabla f(W^t)}}_{\infty}
		 + \norm{B_t} \right)\norm{\hat W^t - W^0}
		 \as 0,
\end{align*}
proving that
\[
	\dist(0, \partial F(\hat W^t)) \as 0.
\]
On the other hand, since $f$ is continuous and $\psi$ is
subdifferentially continuous at $\bar W$ (which implies $F$ is also
subdifferentially contnuous at $\bar W$), $\hat W^t \rightarrow \bar W$,
and that $\nabla f(\hat W^t) + Y_t \as 0 \in \partial F(\bar W)$ (from
\cref{thm:stationary}),
we know that $F(\hat W^t) \as F(\bar W)$ as well.
Therefore, we can apply Lemma~1 of \cite{Lee23a} to conclude that
\cref{eq:identify} indeed holds for some $T_0 < \infty$ almost surely.
\end{proof}

\subsection{Convergence Result for \proxgen}
We next discuss the convergence result for the framework of
\cite{JY20a} with inexactness added.
For consistency, we first use our notations to introduce their
framework, with our
inexactness condition added, in \cref{alg:proxgen}.

\begin{algorithm}[tb]
\LinesNumbered
\DontPrintSemicolon
\caption{\proxgen $(W^0, T, T_2, \{\eta_t\}, \{\rho_t\}, \{c_t\}, \{\epsilon_t\},
\{b_t\}, \delta I)$}
\label{alg:proxgen}

$m_0 \leftarrow 0$

\For{$t=1,\dotsc,T$}
{
	Sample $\xi_t \sim \gD$ with batch size $b_t$

	$ G^t \leftarrow \nabla f_{\xi_t}(W^{t-1})$

	$m_t \gets  \rho_t m_{t-1} + (1 - \rho_t)G^t$

	Construct $P^t$ satisfying $P^t \succeq \delta I$
	
	$\theta_t \gets 1 / \norm{P^t}_2$

	Compute $W^{t+1}$ by roughly solving \cref{eq:proxgen} that
	satisfies \cref{eq:inexact} with $Q_t$ replaced by $\hat Q_t$ and
	$\hat W^t$ replaced by $W^{t+1}$, using \pg$(W^t, W^t, m_t,
	\eta_t^{-1} P^t, \theta_t, T_2, \epsilon_t)$
}
\Output{$W^T$}
\end{algorithm}

In their analysis, \cite{JY20a} made the following four assumptions,
and we will follow these assumptions using the notation
\cref{eq:notation}.
\begin{enumerate}[label=\textbf{(C-$\bf{\arabic*}$)}, ref=\textnormal{(C-$\arabic*$)}, start=1, leftmargin=1.2cm]
	\item The expected loss function $f$ is
		$L$-Lipschitz-continuously-differentiable and lower-bounded
		for some $L \geq 0$.
		\label{con:smooth}
	\item The stochastic gradient $G^t = \nabla
		f_{\xi_t}(W^{t-1})$ is an
		unbiased estimator of $\nabla f(W^{t-1})$ with bounded variance.
		\[
			E_{\xi_t \sim \gD} [ G^t] = \nabla f(W^{t-1}),\quad E_{\xi_t
			\sim D} \left[ \norm{G^t - \nabla f(W^{t-1})}^2 \right]
			\leq \frac{\sigma^2}{b_t}, \quad \forall t\geq 0,
		\]
		where $b_t$ is the batch size of $\xi_t$ and $\sigma \geq 0$
		is a constant.
		\label{con:var}
	\item There are some $\rho_0, \mu \in [0,1)$ and $D, G > 0$ such that
		$\norm{W^{t+1} - W^t} \leq D$, $\norm{G^t} \leq G$, and $\rho_t =
		\rho_0 \mu^{t-1}$ for all $t$.
			\label{con:mild}
	\item There is some $\gamma > 0$ such that $\norm{\eta_t^{-1} P^t}_2 \leq 1 / \gamma <
		\infty$ for all $t$.
		\label{con:mineig}
	\item There is $\delta > 0$ such that \begin{equation}
	P^t \geq \delta,\quad \eta_t \leq \frac{\delta}{3L},\quad \forall t
	\geq 0.
	\label{eq:stepnorm}
\end{equation}
\label{con:maxeig}
\end{enumerate}

\begin{reptheorem}{thm:proxgen}
	For the framework in \cite{JY20a} with the subproblem solved
	approximately by \cref{alg:solver} such that \cref{eq:inexact}
	holds with $\{\epsilon_t\}$ satisfying \cref{eq:eps}.
	Then Theorem~1 of \cite{JY20a} still holds, but with the constants
	$\{Q_i\}$ being also dependent on $\bar \epsilon$.
\end{reptheorem}

\begin{proof}
	The major flow of our proof follows that of \cite{JY20a} but with
	suitable modifications to accommodate the inexactness condition in
	the subproblem solving.

	It is clear from \citep[Lemma~1]{JY20a} that
	\begin{equation}
		\norm{m_t} \leq G,\quad \forall t \geq 0.
		\label{eq:m}
	\end{equation}
By the update rule for $m_t$, \cref{eq:inexact,eq:proxgen}, we have that there is $z_t$ such that
\begin{align*}
0 \in z_t + (1 - \rho_t) g_t + \rho_t m_{t-1} +
{\partial}\psi(W^t) + \frac{1}{\eta_t} (P^t) (W^t - W^{t-1}),
\quad \norm{z_t} \leq \epsilon_t, \quad \forall t \geq 0,
\end{align*}
leading to
\begin{align}
\nabla f(W^t) - z_t - (1 - \rho_t) g_t - \rho_t m_{t-1} -
\frac{1}{\eta_t} (P^t)(W^t - W^{t-1}) \in  {\partial} F(W^t).
\label{eq:opt}
\end{align}
We thus have from \cref{eq:opt} and \ref{con:mineig} that
\begin{align}
	\nonumber
& \dist(0, {\partial}F(W^t))^2 \\
	\nonumber
\leq
~& \Big\|z_t + (1 - \rho_t)g_t - \nabla f(W^t) + \rho_t m_{t-1} + (W^t - W^{t-1}) + \frac{1}{\eta_t} (P^t)(W^t - W^{t-1}) - (W^t - W^{t-1})\Big\|^2 \\
	\nonumber
\leq~& 4\Big\|(1 - \rho_t)g_t - \nabla f(W^t) + \rho_t m_{t-1} + (W^t
- W^{t-1})\Big\|^2 + 4 \epsilon_t^2
+ 4\Big\|\frac{1}{\eta_t} (P^t)(W^t - W^{t-1})\Big\|^2 \\
	\nonumber
\quad &+4\Big\|(W^t - W^{t-1})\Big\|^2 \\
\leq~& 4\underbrace{\Big\|(1 - \rho_t)g_t - \nabla f(W^t) + \rho_t
	m_{t-1} + (W^t - W^{t-1})\Big\|^2}_{T_1} +
	4\Big(\frac{1}{\gamma^2} + 1\Big)\|W^t - W^{t-1}\|^2 + 4
	\epsilon_t^2.
	\label{eq:goal}
\end{align}
We will separately bound the quantities $T_1$ and $\norm{W^t -
	W^{t-1}}^2$ below.

From the subproblem objective requirement in \cref{eq:inexact}, we
get
\begin{equation}
\label{eq:reg}
	\big\langle (1 - \rho_t)g_t + \rho_t m_{t-1}, W^t - W^{t-1}
	\big\rangle + \psi(W^t) + \frac{1}{2\eta_t} \inprod{W^t -
		W^{t-1}}{P^t(W^t - W^{t-1})}\\
		\leq \psi(W^{t-1}).
\end{equation}
From \ref{con:smooth}, we have
\begin{align}
	\label{eq:Lip}
f(W^t) \leq f(W^{t-1}) + \langle \nabla f(W^{t-1}), W^t - W^{t-1} \rangle
+ \frac{L}{2} \|W^t - W^{t-1} \|^2.
\end{align}
Summing \cref{eq:Lip,eq:reg} gives 
\begin{equation}
\label{eqn:core}
	\big\langle (1 - \rho_t) g_t - \nabla f(W^{t-1}) + \rho_t m_{t-1},
	W^t - W^{t-1} \big\rangle + \norm{W^t -
		W^{t-1}}^2_{\frac{P^t}{2\eta_t} - \frac{L}{2} I}
		\\
		\leq F(W^{t-1}) - F(W^t).
\end{equation}
Note that $ \eta_t^{-1} P^t - LI \succeq 0$ from
\cref{eq:stepnorm}, so the second term in \cref{eqn:core} is
nonnegative.
\cref{eqn:core} together with \ref{con:mild} then leads to
\begin{align*}
& \|W^t - W^{t-1}\|_{\frac{P^t}{2\eta_t} - \frac{L}{2}I}^2 \\
\leq
~& F(W^{t-1}) - F(W^t) - \big\langle g_t - \nabla f(W^{t-1}), W^t - W^{t-1} \big\rangle + \langle \rho_t g_t, W^t - W^{t-1} \rangle - \langle \rho_t m_{t-1}, W^t - W^{t-1} \rangle \\
\leq~& F(W^{t-1}) - F(W^t) + \frac{1}{2L} \|g_t - \nabla f(W^{t-1})\|^2 + \frac{L}{2} \|W^t - W^{t-1}\|^2 + \frac{\rho_t^2}{2L}\|g_t\|^2 + \frac{L}{2} \|W^t - W^{t-1}\|^2 \\
& \quad + \rho_t \|m_{t-1}\| \|W^t - W^{t-1}\| \\
\leq~& F(W^{t-1}) - F(W^t) +
\frac{1}{2L} \|g_t - \nabla f(W^{t-1})\|^2 +
+ L \|W^t - W^{t-1}\|^2 +
\frac{\rho_0^2 \mu^{2(t-1)}G^2}{2L} + \rho_0 \mu^{t-1} DG.
\end{align*}
Summing it over $t=1,2,\dotsc,T$ and utilizing the assumption that the step
sizes are non-increasing then give
\begin{align*}
\Big(\frac{\delta}{2\eta_0} - \frac{3}{2}L\Big)
\sum\limits_{t=1}^{T} \|W^t - W^{t-1}\|^2 \leq \Delta + C_1 + \frac{1}{2L} \sum\limits_{t=1}^{T} \|g_t - \nabla f(W^{t-1})\|^2,
\end{align*}
where
\[
	\Delta \coloneqq F(W^0) - \min_{W} F(W),\quad C_1 \coloneqq
	\frac{\rho_0 DG}{1 - \mu} + \frac{\rho_0^2G^2}{2L(1 - \mu^2)}.
\]
From the inequality above, we obtain
\begin{align}\label{eqn:distance_bound}
\sum\limits_{t=1}^{T} \|W^t - W^{t-1}\|^2 \leq H_1 + H_2 \sum\limits_{t=1}^{T}\|g_t - \nabla f(W^{t-1})\|^2
\end{align}
for some constants $H_1, H_2$ depending on $L, \Delta, \delta, \eta_0$, and $C_1$.
From \cref{eqn:core}, we also have
\begin{align*}
& \Big\langle (1 - \rho_t) g_t - \nabla f(W^t) + \rho_t m_{t-1}, W^t - W^{t-1} \Big\rangle \\
\leq~& F(W^{t-1}) - F(W^t) - \big\langle \nabla f(W^t) - \nabla f(W^{t-1}), W^t - W^{t-1} \big\rangle - \big\|W^t - W^{t-1}\|_{\frac{1}{2\eta_t}(P^t) - \frac{L}{2} I}^2 \\
\leq~& F(W^{t-1}) - F(W^t) - \big\langle \nabla f(W^t) - \nabla
f(W^{t-1}), W^t - W^{t-1} \big\rangle.
\end{align*}
Therefore, we obtain
\begin{align}
	\nonumber
T_1 & = \|(1 - \rho_t)g_t - \nabla f(W^t) + \rho_t m_{t-1}\|^2 + \|W^t - W^{t-1}\|^2
+ 2\Big\langle (1 - \rho_t) g_t - \nabla f(W^t) + \rho_t m_{t-1}, W^t - W^{t-1} \Big\rangle \\
	\nonumber
& \leq \|(1 - \rho_t)g_t - \nabla f(W^{t-1}) + \nabla f(W^{t-1}) - \nabla f(W^t) + \rho_t m_{t-1}\|^2 + \|W^t - W^{t-1}\|^2 \\
	\nonumber
& \quad + 2\left( F(W^{t-1}) - F(W^t) - \big\langle \nabla f(W^t) - \nabla
f(W^{t-1}), W^t - W^{t-1} \big\rangle\right)\\
	\nonumber
& \leq 4\|g_t - \nabla f(W^{t-1})\|^2 + 4L^2\|W^t - W^{t-1}\|^2 +
4\rho_t^2(\|m_{t-1}\|^2 + \|g_t \|^2) + \|W^t - W^{t-1}\|^2 \\
	\nonumber
& \quad +2\left(F(W^{t-1}) - F(W^t) + L\|W^t - W^{t-1}\|^2\right)\\
& \leq 2\left(F(W^{t-1}) - F(W^t)\right) + 8\rho_0^2 \mu^{2(t-1)} G^2
 + \Big(1 + 2L + 4L^2\Big) \|W^t - W^{t-1}\|^2 + 4 \|g_t - \nabla
 f(W^{t-1})\|^2.
	\label{eq:T1}
\end{align}
Let $C_2 \coloneqq 2+2L+4L^2+\gamma^{-2}$ and
insert \cref{eq:T1} into \cref{eq:goal}, we get
\begin{align}
	\nonumber
& \dist(0, {\partial}F(W^t))^2 \\
\leq~& 4\Bigg(2\left(F(W^{t-1}) - F(W^t)\right) + 8\rho_0^2 \mu^{2(t-1)} G^2 +
{C_2}\|W^t - W^{t-1}\|^2 + 4\|g_t - \nabla f(W^{t-1})\|^2 +
\epsilon_t^2 \Bigg).
\label{eq:single}
\end{align}
Therefore, we have from \cref{eq:single,eq:eps} and
\ref{con:var} that
\begin{align*}
	&~\E_{a,\xi_1,\dotsc,\xi_T}[\dist(0, {\partial}F(W^a))^2] \\
	\leq
	&~ \frac{1}{T} \sum_{t=1}^{T} \E\Big[\big\|(1 -
	\rho_t)g_t - \nabla f(W^t) + z_t + \rho_t m_{t-1} + \frac{1}{\eta_t} (P^t)(W^t - W^{t-1})\big\|^2\Big] \\
\leq
&~\frac{4}{T} \Big(2\Delta + \frac{8\rho_0^2 G^2}{1 - \mu^2} +
4\sum_{t=1}^{T} \E \|g_t - \nabla f(W^{t-1})\|^2 + C_2
\sum_{t=1}^{T} \E\|W^t - W^{t-1}\|^2 + \sum_{t=1}^T \epsilon_t^2 \Big) \\
\leq&~ \frac{4}{T}\Big(2\Delta + \frac{8\rho_0^2 G^2}{1 - \mu^2} +
4\sigma^2 \sum_{t=1}^{T} \frac{1}{b_t} + C_2 (H_1 + H_2 \sigma^2 \sum_{t=1}^{T}\frac{1}{b_t}
)+ \bar \epsilon \Big) \\
\leq &~\frac{Q_1}{T} \sum_{t=1}^{T} \frac{1}{b_t}  +
\frac{Q_2\Delta}{T} + \frac{Q_3}{T},
\end{align*}
for some constants $Q_1,Q_2,Q_3$
dependent on $\{\eta_0, \delta, \Delta, L, D, G, \rho_0, \mu,
\gamma,\bar \epsilon\}$, but not on $T$.
This proves our theorem.
\end{proof}

\section{Additional Experiments for Computer Vision}
\label{app:moreexp}
In this section, we compare \ramda with other methods on image
classification with smaller datasets.
They are:
\begin{enumerate}
    \item Logistic regression (neural network with no hidden layer) with the MNIST dataset \citep{lecun1998gradient}.
    \item A modified VGG19 \citep{simonyan2014very} with the CIFAR10 dataset \citep{krizhevsky2009learning}.
    \item The same VGG19 with the CIFAR100 dataset \citep{krizhevsky2009learning}.
    \item A modified ResNet50 \citep{he2016deep} with the CIFAR10 dataset.
    \item The same ResNet50 with the CIFAR100 dataset.
\end{enumerate}

The results are shown in \cref{tbl:cv2}. 
In the logistic regression problem, we only perform a single run, with
the initial point being the origin, as it is a convex problem.
Moreover, in this problem, when dealing with \proxssi, \proxgen, and \proxsgd
whose sparsity levels are highly unstable over iterations,
we report their highest weighted group sparsity over all epochs, but for all other problems, we report the group sparsity level of the final output.

Experiments in this subsection show that \ramda might sometimes
perform worse than existing methods on smaller problems like
CIFAR10/100.
Fortunately, for such smaller problems, the training cost is not very
significant, and one can afford to try more algorithms.

\begin{table}[tb]
\caption{Group sparsity and validation accuracy of different methods
	  on image classification with smaller datasets.}
\label{tbl:cv2}
\begin{center}
\renewrobustcmd{\bfseries}{\fontseries{b}\selectfont}
\renewrobustcmd{\boldmath}{}
\newrobustcmd{\B}{\bfseries}
\begin{center}
\begin{tabular}{@{}llrr@{}}
\hline
Algorithm & Validation accuracy & Group sparsity \\
\hline
\multicolumn{3}{c}{Logistic Regression/MNIST} \\
\hline
\proxsgd & 91.31\% & 39.29\% \\
\proxssi & 91.31\% & 39.92\% \\
\proxgen & 91.31\% & 39.92\% \\
\rmda & 91.34\% & 57.02\% \\
\ramda & \B{91.35\%} & \B{57.40\%} \\
\hline
\multicolumn{3}{c}{VGG19/CIFAR10} \\
\hline
\msgd & 93.95 $\pm$ 0.14\% & \multicolumn{1}{c}{-} \\
\hline
\hline
\proxsgd & 92.82 $\pm$ 0.09\% & 82.76 $\pm$ 5.42\% \\
\proxssi & 92.81 $\pm$ 0.15\% & 88.40 $\pm$ 0.23\% \\
\proxgen & 92.83 $\pm$ 0.05\% & 86.64 $\pm$ 0.12\% \\
\rmda & \B{93.13 $\pm$ 0.10\%} & \B{90.22 $\pm$ 0.06\%} \\
\ramda & 92.89 $\pm$ 0.13\% & 86.31 $\pm$ 0.31\% \\
\hline
\multicolumn{3}{c}{VGG19/CIFAR100} \\
\hline
\msgd & 74.07 $\pm$ 0.05\% & \multicolumn{1}{c}{-} \\
\hline
\hline
\proxsgd & \B{71.96 $\pm$ 0.15\%} & 72.34 $\pm$ 11.9\% \\
\proxssi & 67.29 $\pm$ 0.06\% & 78.58 $\pm$ 0.34\% \\
\proxgen & 68.13 $\pm$ 0.36\% & 75.46 $\pm$ 0.17\% \\
\rmda & \B{71.96 $\pm$ 0.31\%} & \B{80.88 $\pm$ 0.11\%} \\
\ramda & 70.47 $\pm$ 0.25\% & 65.19 $\pm$ 0.77\% \\
\hline
\multicolumn{3}{c}{ResNet50/CIFAR10} \\
\hline
\msgd & 95.54 $\pm$ 0.19\% & \multicolumn{1}{c}{-} \\
\hline
\hline
\proxsgd & 92.36 $\pm$ 0.05\% & 82.18 $\pm$ 2.67\% \\
\proxssi & 94.04 $\pm$ 0.12\% & 83.67 $\pm$ 0.63\% \\
\proxgen & 94.07 $\pm$ 0.12\% & 80.45 $\pm$ 0.45\% \\
\rmda & \B{95.11 $\pm$ 0.11\%} & \B{85.64 $\pm$ 0.12\%} \\
\ramda & 93.85 $\pm$ 0.10\% & 81.99 $\pm$ 1.26\% \\
\hline
\multicolumn{3}{c}{ResNet50/CIFAR100} \\
\hline
\msgd & 79.49 $\pm$ 0.49\% & \multicolumn{1}{c}{-} \\
\hline
\hline
\proxsgd & 74.54 $\pm$ 0.58\% & 49.29 $\pm$ 5.91\% \\
\proxssi & 73.65 $\pm$ 0.39\% & 70.38 $\pm$ 0.74\% \\
\proxgen & 73.63 $\pm$ 0.43\% & 65.51 $\pm$ 3.58\% \\
\rmda & \B{75.62 $\pm$ 0.19\%} & \B{79.97 $\pm$ 0.27\%} \\
\ramda & 69.23 $\pm$ 0.86\% & 68.65 $\pm$ 1.83\% \\
\end{tabular}
\end{center}
\end{center}
\end{table}

\section{Plots of Sparsity Level and Validation Accuracy over Epochs}
\label{app:plot}
\begin{figure}
\begin{center}
\begin{subfigure}[b]{0.47\textwidth}
	\centering
	\includegraphics[width=\textwidth]{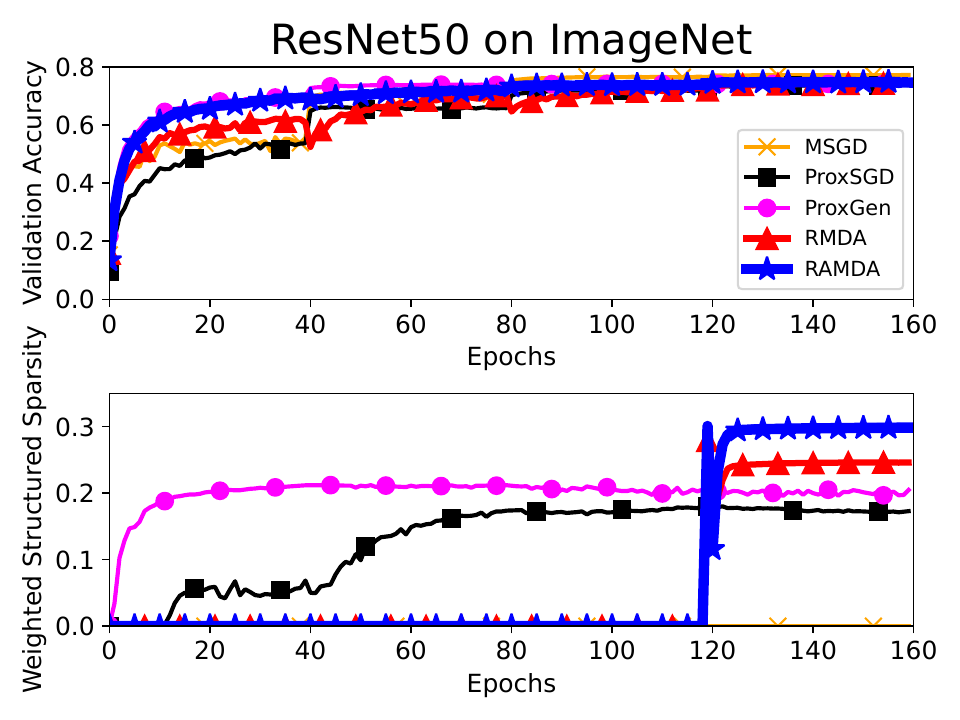}
	\caption{ResNet50 on ImageNet}
\end{subfigure}
\begin{subfigure}[b]{0.47\textwidth}
	\centering
	\includegraphics[width=\textwidth]{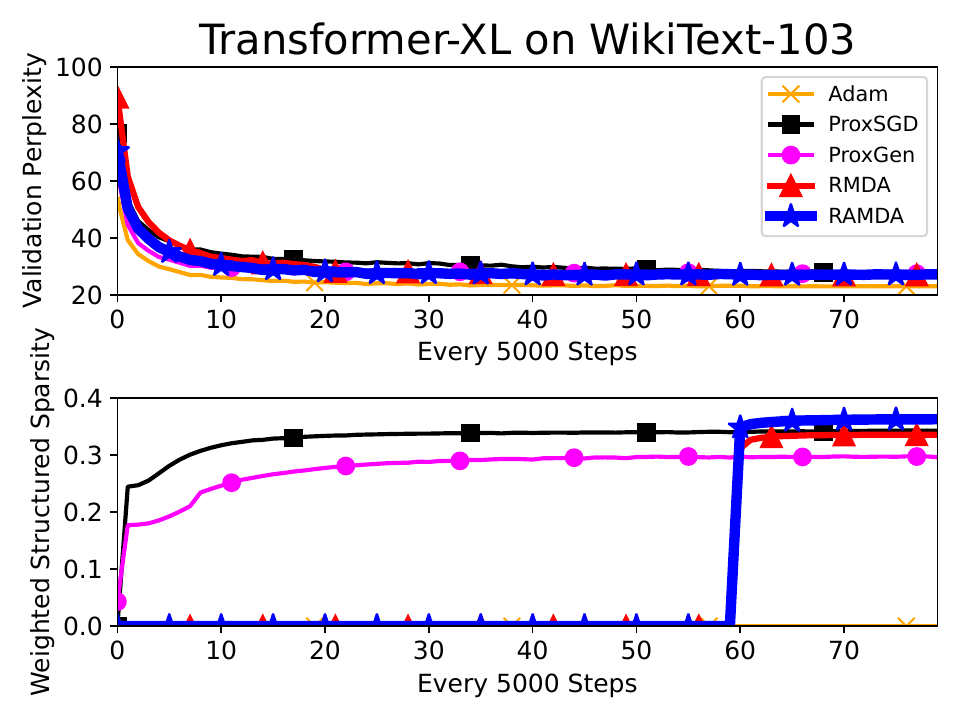}
	\caption{Transformer-XL on WikiText-103}
\end{subfigure}
\begin{subfigure}[b]{0.47\textwidth}
	\centering
	\includegraphics[width=\textwidth]{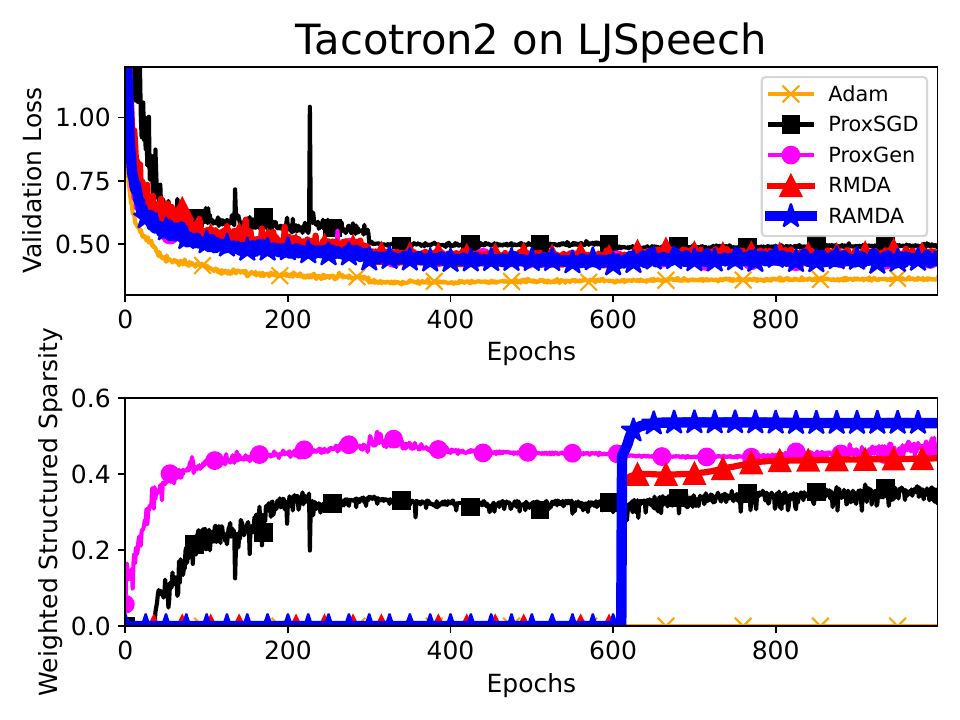}
	\caption{Tacotron2 on LJSpeech}
\end{subfigure}
\begin{subfigure}[b]{0.47\textwidth}
	\centering
	\includegraphics[width=\textwidth]{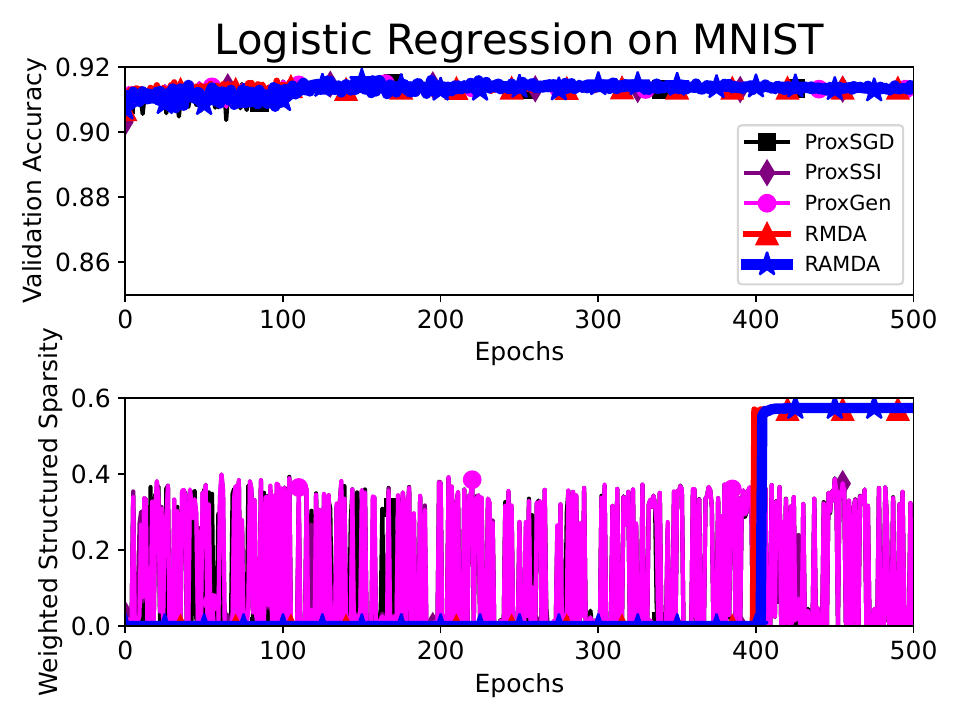}
	\caption{Logistic Regression on MNIST}
\end{subfigure}
\begin{subfigure}[b]{0.47\textwidth}
	\centering
	\includegraphics[width=\textwidth]{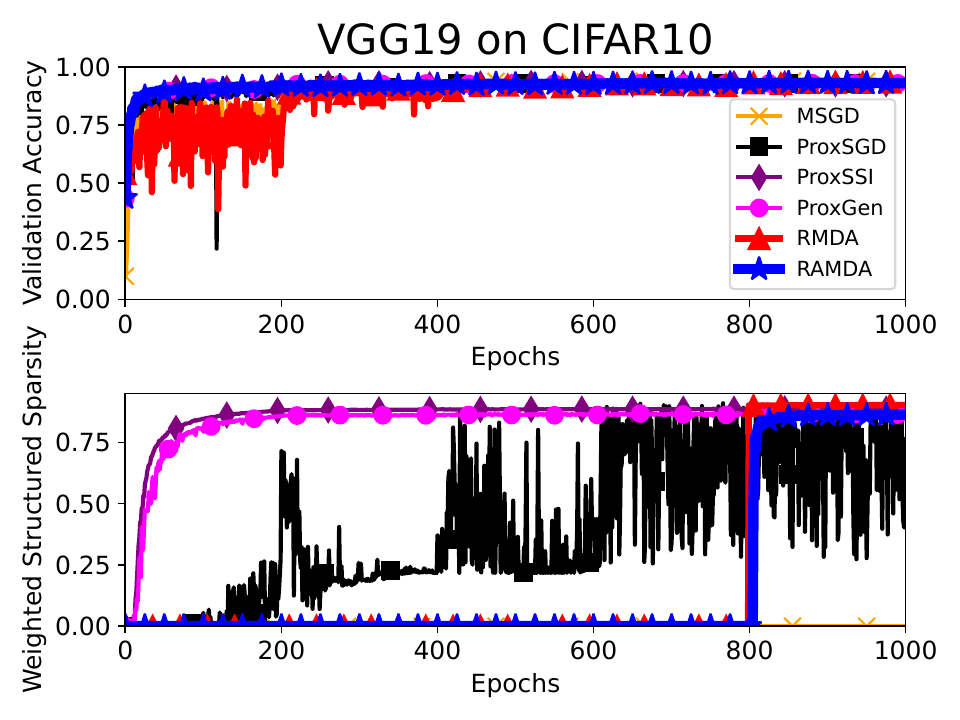}
	\caption{VGG19 on CIFAR10}
\end{subfigure}
\begin{subfigure}[b]{0.47\textwidth}
	\centering
	\includegraphics[width=\textwidth]{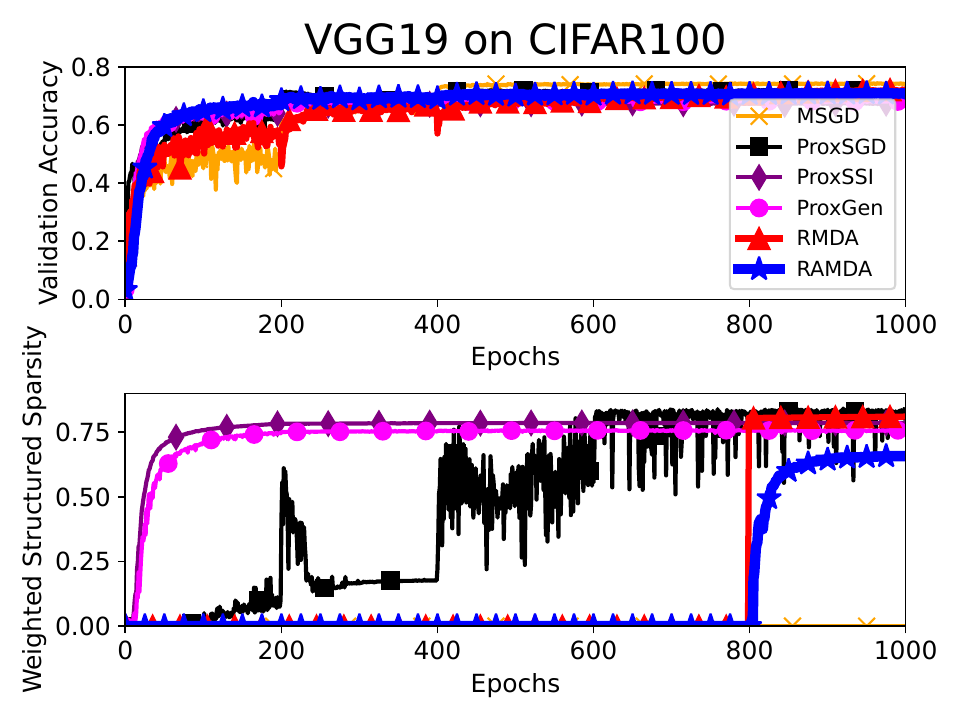}
	\caption{VGG19 on CIFAR100}
\end{subfigure}
\begin{subfigure}[b]{0.47\textwidth}
	\centering
	\includegraphics[width=\textwidth]{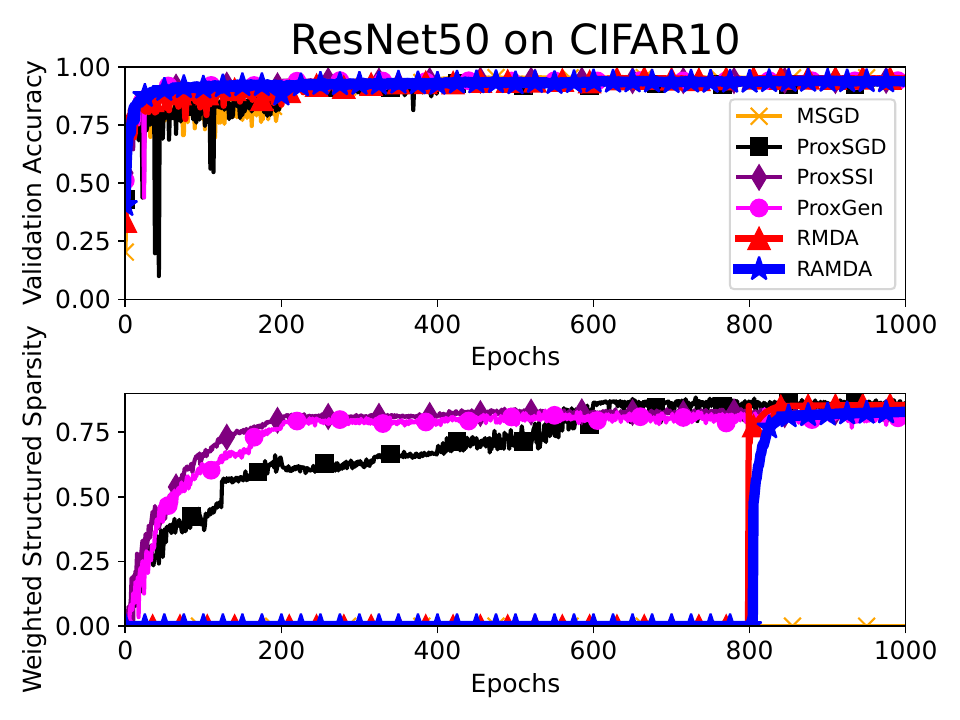}
	\caption{ResNet50 on CIFAR10}
\end{subfigure}
\begin{subfigure}[b]{0.47\textwidth}
	\centering
	\includegraphics[width=\textwidth]{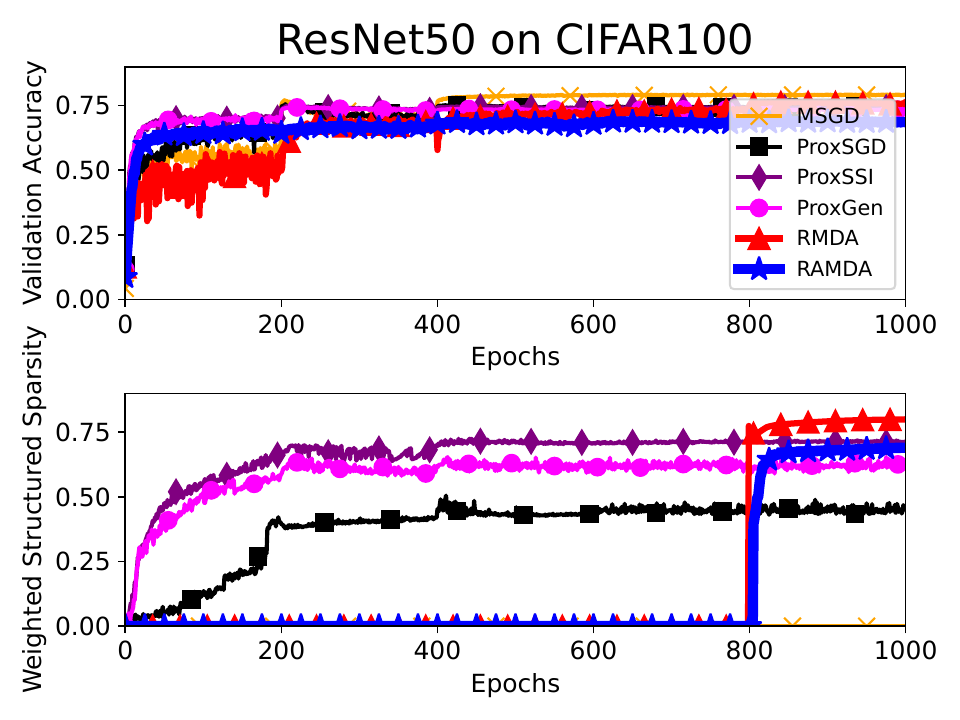}
	\caption{ResNet50 on CIFAR100}
\end{subfigure}
\end{center}
\caption{Group sparsity level and validation prediction performance v.s
epochs.
In the plot for Transformer-XL, one step processes ten batches, and
for our batch size of 64, one epoch consists of 8,401 batches.}
\label{fig:all}
\end{figure}

\begin{figure}
\begin{center}
\begin{subfigure}[b]{0.47\textwidth}
	\centering
	\includegraphics[width=\textwidth]{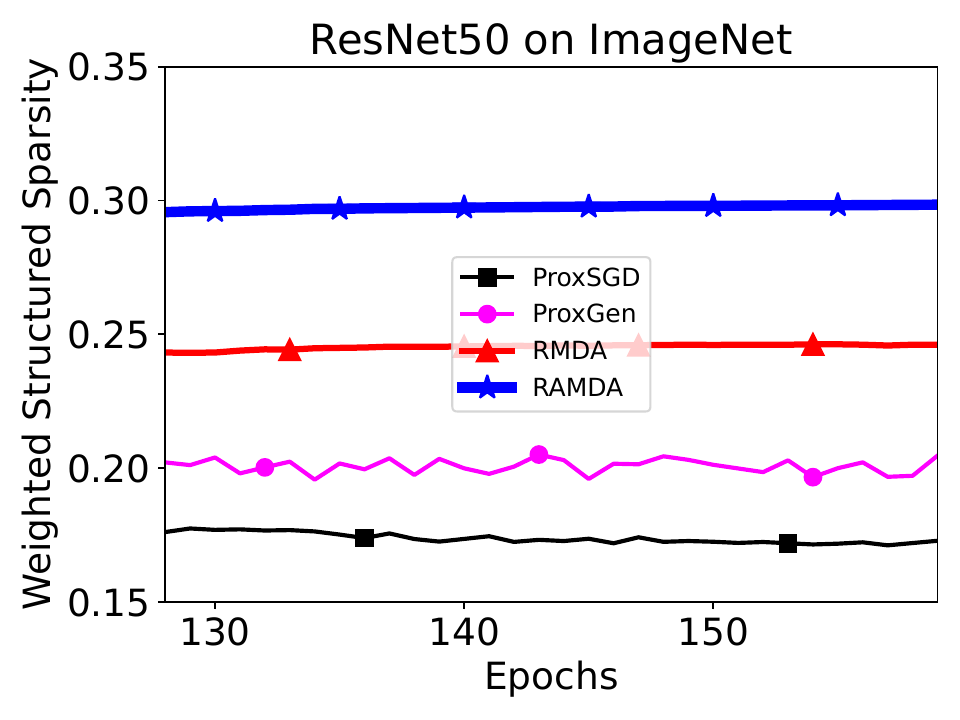}
	\caption{ResNet50 on ImageNet}
\end{subfigure}
\begin{subfigure}[b]{0.47\textwidth}
	\centering
	\includegraphics[width=\textwidth]{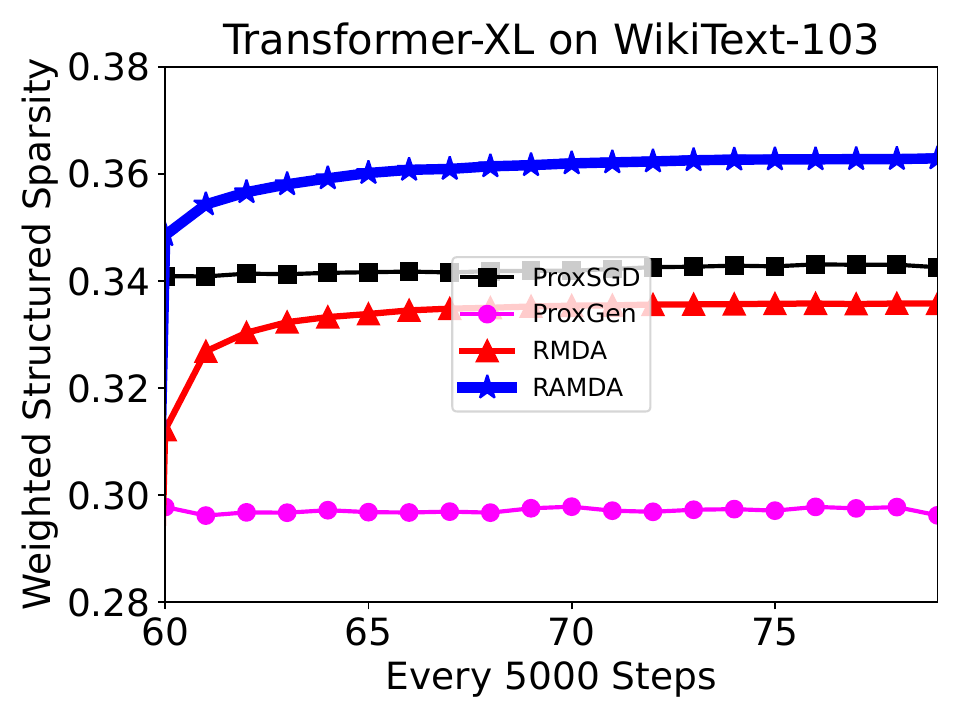}
	\caption{Transformer-XL on WikiText-103}
\end{subfigure}
\begin{subfigure}[b]{0.47\textwidth}
	\centering
	\includegraphics[width=\textwidth]{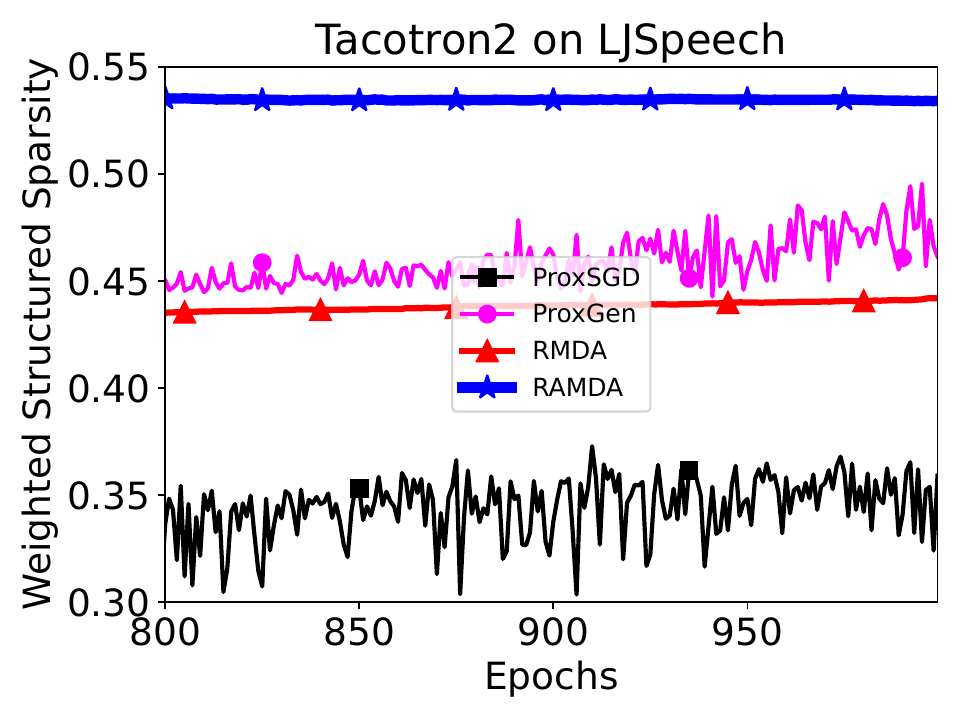}
	\caption{Tacotron2 on LJSpeech}
\end{subfigure}
\begin{subfigure}[b]{0.47\textwidth}
	\centering
	\includegraphics[width=\textwidth]{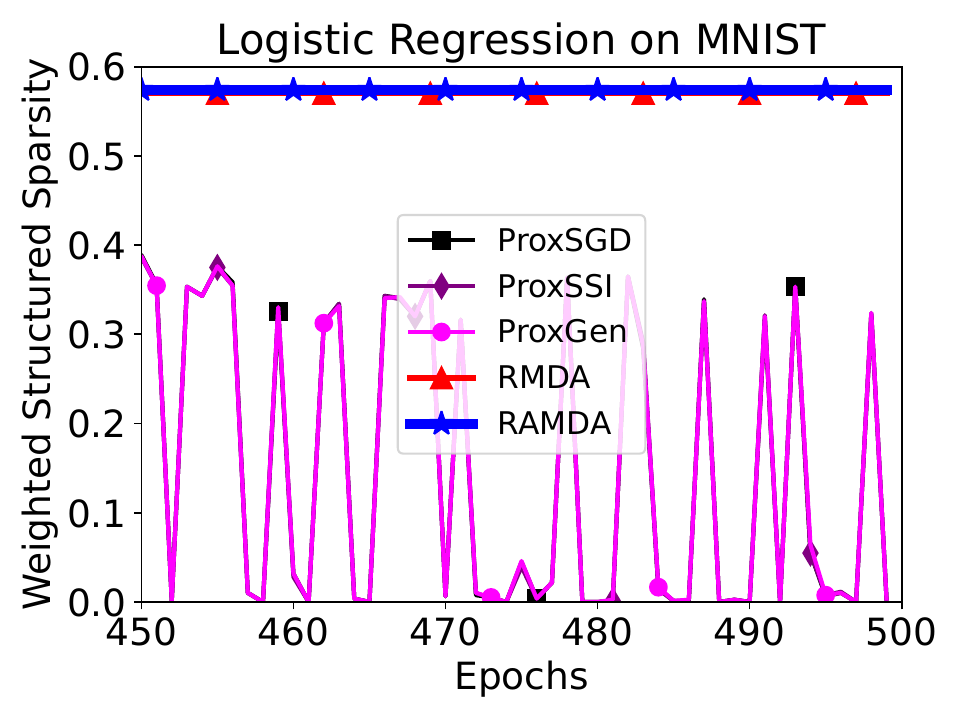}
	\caption{Logistic Regression on MNIST}
\end{subfigure}
\begin{subfigure}[b]{0.47\textwidth}
	\centering
	\includegraphics[width=\textwidth]{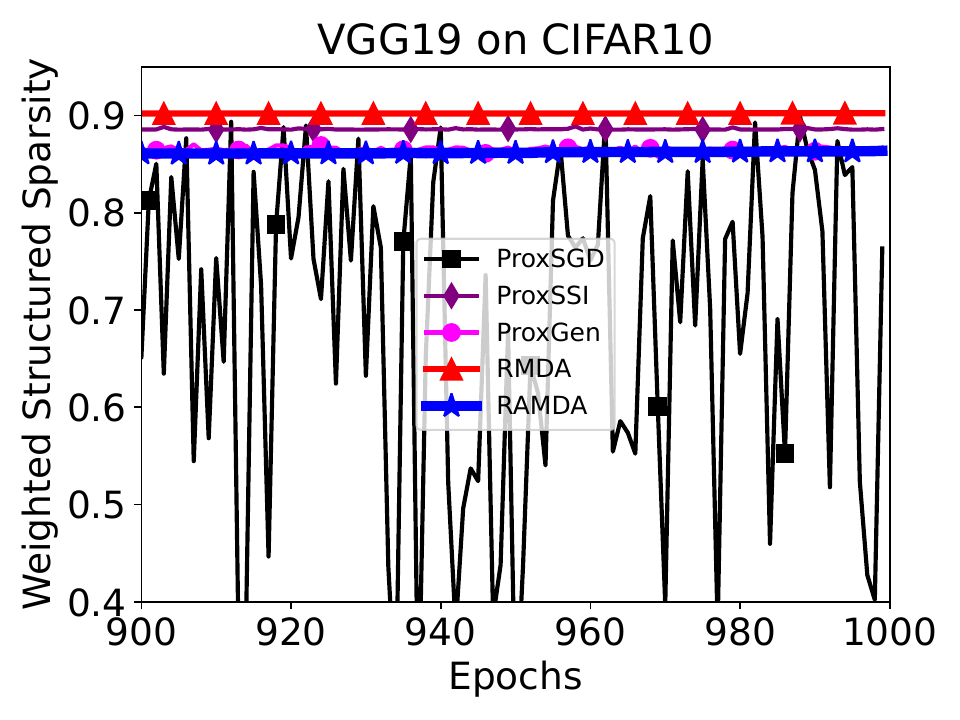}
	\caption{VGG19 on CIFAR10}
\end{subfigure}
\begin{subfigure}[b]{0.47\textwidth}
	\centering
	\includegraphics[width=\textwidth]{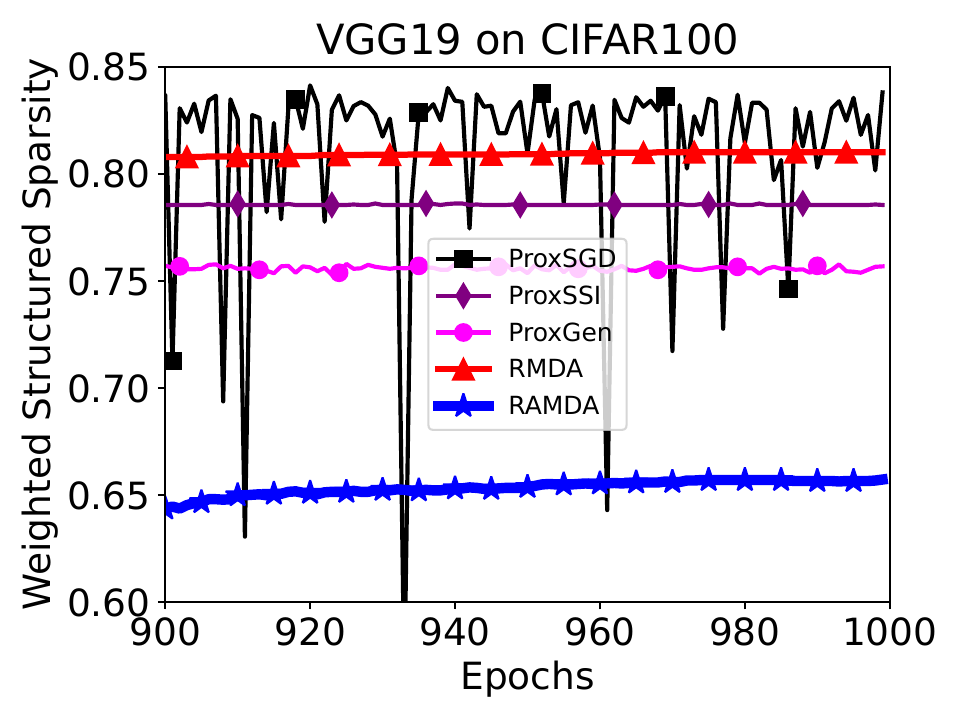}
	\caption{VGG19 on CIFAR100}
\end{subfigure}
\begin{subfigure}[b]{0.47\textwidth}
	\centering
	\includegraphics[width=\textwidth]{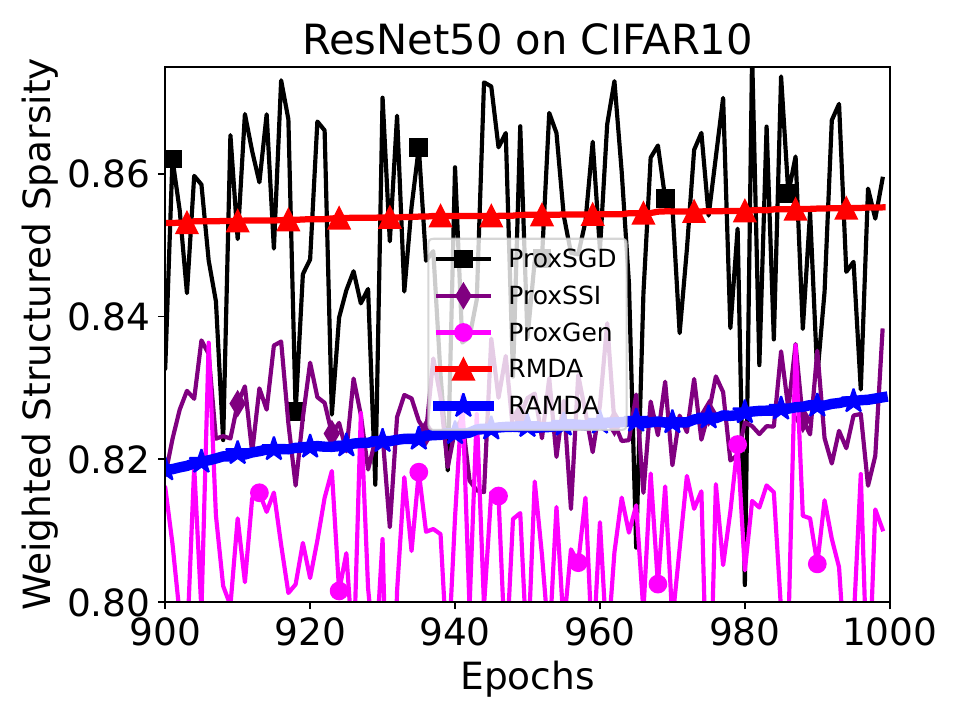}
	\caption{ResNet50 on CIFAR10}
\end{subfigure}
\begin{subfigure}[b]{0.47\textwidth}
	\centering
	\includegraphics[width=\textwidth]{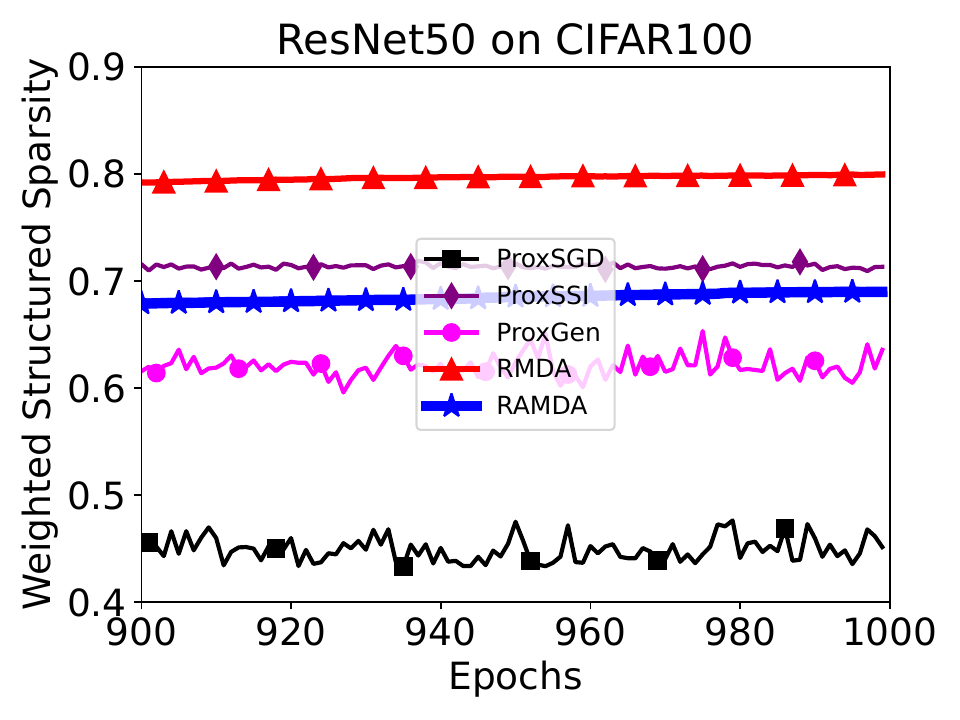}
	\caption{ResNet50 on CIFAR100}
\end{subfigure}
\end{center}
\caption{Group sparsity level at the last epochs.}
\label{fig:sparsity}
\end{figure}

We provide in \cref{fig:all} the plots of predictive ability and
structured sparsity over epochs of some representative experiments we
have conducted.
These experiments are:
\begin{enumerate}
    \item ResNet50 with the ILSVRC 2012 ImageNet dataset.
    \item Transformer-XL with the WikiText-103 dataset.
    \item Tacotron2 with the LJSpeech dataset.
    \item Logistic Regression with the MNIST dataset.
    \item A modified VGG19  with the CIFAR10 dataset.
    \item The same VGG19 with the CIFAR100 dataset.
    \item A modified ResNet50 with the CIFAR10 dataset.
    \item The same ResNet50 with the CIFAR100 dataset.
\end{enumerate}
In the plot for Transformer-XL, one step processes ten batches, and
for our batch size of 64, one epoch consists of 8,401 batches.
We further observe in the zoomed-in sparsity plots in
\cref{fig:sparsity} that the sparsity level of \ramda is stable
at the final epochs.
These plots corroborates our theory that
\ramda is indeed capable of manifold identification while achieving competitive prediction performance.
On the other hand, in the absence of manifold identification
guarantees, the sparsity levels of \proxsgd, \proxssi and \proxgen
exhibit oscillations that are sometimes drastic.
We note that for the largest problems Tacotron2 and Transformer-XL,
the sparsity levels of \ramda are still gradually increasing even at
the final epochs.
This suggests that if we are willing to run the algorithm for longer,
it is possible that the structured sparsity level could be further improved.

\section{Experiment with Nuclear-norm Regularization}
\label{app:nuc}
We further conduct some preliminary experiments with a different
regularizer to showcase that the proposed \ramda can be applied to
structures beyond sparsity.
We consider the structure such that each layer of the neural network
is low-rank, induced by imposing one nuclear-norm regularizer per
layer individually by treating each layer as a matrix.
Given a matrix $X \in \R^{m \times n}$ of rank $r \leq \min\{m,n\}$
with its singular value decomposition (SVD) $X = U \Sigma V^{\top}$,
where $U \in \R^{m\times r}$, $V \in \R^{n \times r}$ are orthogonal
and the positive definite diagonal matrix $\Sigma \in \R^{r \times r}$
represents the nonzero singular values of $X$,
the nuclear norm of $X$ is computed by
\[
	\norm{X}_* = \sum_{i=1}^r \Sigma_{i,i},
\]
and the corresponding proximal operator for $\lambda > 0$ is
\[
	\prox_{\lambda \norm{\cdot}_*}(X) = U \hat \Sigma V^{\top}, \text{
	where } \hat \Sigma_{i,i} = \max\{0, \Sigma_{i,i} - \lambda\},
	\quad i=1,\dotsc,r.
\]
Given a point $X^*$ with rank $r^*$, the active manifold of the
nuclear norm at $X^*$ is
\[
	\mathcal{M}(X^*) = \{Y \mid \text{rank}(Y) = r^*\}.
\]
Using low-rankness to condense neural networks is itself an
interesting research topic, but conducting full SVDs could be rather
time-consuming, so applying this structure to larger problems is
challenging but potentially useful.
How to exploit this structure for prediction acceleration and to make
the training more efficient, possibly using iterative methods to compute
approximate SVDs, is an interesting topic we plan to investigate in
the near future.
Instead, the purpose of the preliminary experiment here is merely for
showing that our method is also applicable to other structures.

We first consider training a simple neural network with six
fully-connected layers using the FashionMNIST dataset
\citep{xiao2017fashion}.
Since this is a rather small-scale problem and this is a image
classification problem, we do not expect \ramda to outperform
non-adaptive methods, especially the \rmda method that is also able to
identify the active manifold.
The goal of this experiment is just to demonstrate the possibilities of
structures beyond sparsity.
The results are shown in \cref{tbl:nuclear}.
As we have anticipated, \ramda is indeed slightly worse than \rmda
regarding the low-rank level and the prediction accuracy, but it is
still competitive and outperforms \proxgen and \proxsgd.
This exemplifies the potential of \ramda as well as \rmda for training
neural networks with other useful structures.

We also conduct an experiment on pretraining a modified vision
transformer model \citep{Liu21a} for masked image modeling
\citep{XieZC22a} using the CIFAR10 dataset.
Following the standard practice of this task, we select the model
that gives the lowest validation loss among the last 50 epochs as the
final output.
The results are shown in \cref{tbl:vit}.
We can see that \ramda attains the lowest validation loss and has a
low-rank level almost identical to that of \rmda.
On the other hand, \proxsgd and \proxgen have worse low-rank levels.

\begin{table}[tb]
\caption{Low-rank level and validation accuracy of different methods
	  on training a six-layer fully-connected neural network with the
  FashionMNIST dataset for image classification.}
\label{tbl:nuclear}
\begin{center}
\renewrobustcmd{\bfseries}{\fontseries{b}\selectfont}
\renewrobustcmd{\boldmath}{}
\newrobustcmd{\B}{\bfseries}
\begin{center}
\begin{tabular}{@{}lrr@{}}
\hline
Algorithm & Validation accuracy & Low-rank level\\
\hline
\msgd & 89.95 $\pm$ 0.29\% &  \multicolumn{1}{c}{-} \\
\hline
\hline
\proxsgd & 87.54 $\pm$ 0.52\% & 78.00 $\pm$ 0.77\% \\
\proxgen & 86.66 $\pm$ 0.33\% & 87.46 $\pm$ 4.19\% \\
\rmda & \B{88.19 $\pm$ 0.23}\% & \B{91.88 $\pm$ 0.12}\% \\
\ramda & 87.99 $\pm$ 0.24\% & 89.59 $\pm$ 0.42\%
\end{tabular}
\end{center}
\end{center}
\end{table}

\begin{table}[tb]
\caption{Low-rank level and validation loss of different methods
	  on pretraining a modified vision transformer model using the
  CIFAR10 dataset for masked image modeling.}
\label{tbl:vit}
\begin{center}
\renewrobustcmd{\bfseries}{\fontseries{b}\selectfont}
\renewrobustcmd{\boldmath}{}
\newrobustcmd{\B}{\bfseries}
\begin{center}
\begin{tabular}{@{}lrr@{}}
\hline
Algorithm & Validation loss & Low-rank level\\
\hline
\adamw &  0.0865 $\pm$ 0.0001 & \multicolumn{1}{c}{-} \\
\hline
\hline
\proxsgd & 0.1042 $\pm$ 0.0003  & 82.60 $\pm$ 0.34\% \\
\proxgen & 0.1120 $\pm$ 0.0019 & 82.64 $\pm$ 2.47\% \\
\rmda & 0.1054 $\pm$ 0.0031 & \B{86.23 $\pm$ 0.41}\% \\
\ramda & \B{0.1035 $\pm$ 0.0016}  & \B{86.20 $\pm$ 0.35}\%
\end{tabular}
\end{center}
\end{center}
\end{table}

\end{document}